\documentclass{article} 
\usepackage{iclr2021_conference,times}

\usepackage{amsmath,amsfonts,bm}

\def\eqref#1{equation~\ref{#1}}

\def\1{\bm{1}}

\DeclareMathAlphabet{\mathsfit}{\encodingdefault}{\sfdefault}{m}{sl}
\SetMathAlphabet{\mathsfit}{bold}{\encodingdefault}{\sfdefault}{bx}{n}

\usepackage{hyperref}
\usepackage{url}

\usepackage{graphicx}
\usepackage{amsmath}
\usepackage{amssymb}
\usepackage{gensymb}

\usepackage{amssymb}
\usepackage{xcolor}

\usepackage{subcaption}

\usepackage{kotex}
\usepackage{color}
\usepackage{array}
\usepackage{hyperref}

\usepackage{bm}
\usepackage{nth}
\usepackage{setspace}

\newcolumntype{C}[1]{>{\centering\let\newline\\\arraybackslash\hspace{0pt}}m{#1}}
\newcolumntype{L}[1]{>{\let\newline\\\arraybackslash\hspace{0pt}}m{#1}}

\usepackage{rotating} 

\usepackage{amsmath}

\usepackage{booktabs}
\usepackage{listings}
\usepackage{algorithm}
\usepackage{algorithmic}
\usepackage{mathtools, nccmath}

\usepackage[capitalize]{cleveref}
\crefname{section}{Sec.}{Secs.}
\Crefname{section}{Section}{Sections}
\Crefname{table}{Table}{Tables}
\crefname{table}{Tab.}{Tabs.}

\title{The Effects of Mixed Sample Data Augmentation are Class Dependent}

\author{Haeil Lee, Hansang Lee\thanks{Haeil Lee and Hansang Lee contribute equally.}, and Junmo Kim \\
School of Electrical Engineering\\
Korea Advanced Institute of Science and Technology\\
Daehark 291, Yuseonggu, Daejeon 34141, Republic of Korea\\
\texttt{\{haeil.lee,hansanglee,junmo.kim\}@kaist.ac.kr}\\
}

\iclrfinalcopy 
\begin{document}

\maketitle

\begin{abstract}

Mixed Sample Data Augmentation (MSDA) techniques, such as Mixup, CutMix, and PuzzleMix, have been widely acknowledged for enhancing performance in a variety of tasks. A previous study reported the class dependency of traditional data augmentation (DA), where certain classes benefit disproportionately compared to others. This paper reveals a class dependent effect of MSDA, where some classes experience improved performance while others experience degraded performance. This research addresses the issue of class dependency in MSDA and proposes an algorithm to mitigate it. The approach involves training on a mixture of MSDA and non-MSDA data, which not only mitigates the negative impact on the affected classes, but also improves overall accuracy. Furthermore, we provide in-depth analysis and discussion of why MSDA introduced class dependencies and which classes are most likely to have them.

\end{abstract}    
\section{Introduction}
\label{sec:intro}

Data augmentation is a crucial element in deep learning for various tasks. 
Recently, mixed sample data augmentation (MSDA) has emerged as a popular technique, which creates new training data by combining multiple images and labels. 
Since Zhang~\textit{et al.} introduced the first MSDA method, Mixup~\citep{mixup}, which generates new data through a linear combination of two different samples, many variants of MSDA with various combination techniques have been proposed~\citep{cutmix,manifoldMixup,resizemix,puzzlemix,saliencymix,automix,supermix,coMixup}.
Despite its simple concept and computation, MSDA has shown significant performance improvement in various tasks, including image recognition~\citep{cutmix,resizemix,puzzlemix,saliencymix,supermix}, semantic segmentation~\citep{msdaseg1,msdaseg2,msdaseg3,msdaseg4,msdaseg5}, natural language processing~\citep{msdanlp1,msdanlp3,msdanlp3,msdanlp4,msdanlp5,msdanlp6}, video processing~\citep{msdavideo1,msdavideo2,msdavideo3,msdavideo4}, and medical image analysis~\citep{msdamed1,msdamed2,msdamed3,msdamed4,msdamed5,msdamed6,msdamed7}.

\begin{figure}[t!]
 \centering
 \includegraphics[width=0.7\linewidth]{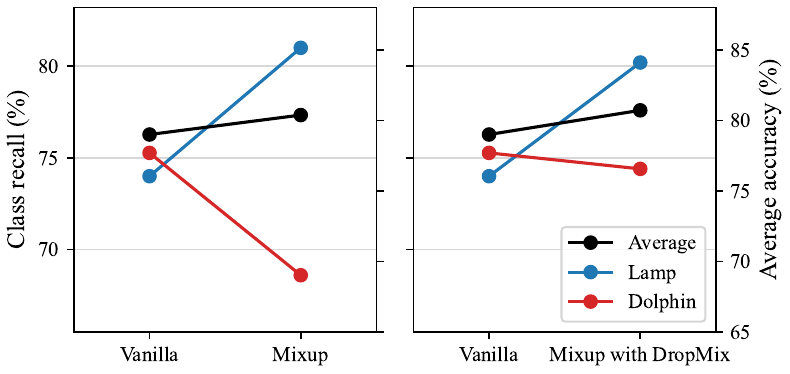}
 \caption{\textbf{Overview.} 
 The effect of MSDA is class dependent.
 Five PreActResNet50 models are trained and averaged on CIFAR-100 dataset with no MSDA, Mixup, and Mixup with the proposed DropMix.
 The left graph shows that while Mixup improves the average accuracy and recall for the ``Lamp'' class, but it reduces recall for the ``Dolphin'' class. 
 This observation led us to address the issue of measuring and reducing class dependency in MSDA effects. 
 Our goal is to minimize the performance degradation in classes such as ``Dolphin'' even if it sacrifices some improvement in classes like ``Lamp.'' 
 The right graph shows that DropMix method has a slight decline in ``Lamp'' performance compared to Mixup but also a much smaller decrease in ``Dolphin'' performance compared to Mixup.
 Additionally, the DropMix method achieves a better overall average accuracy than Mixup.
 }
 \label{fig:overview}
\end{figure}

Several studies~\citep{Mixupcalib1,Mixupcalib2,Mixupcalib3,Mixuprobust,Mixupimbal,Mixupdatadepend,Mixupcutmix} have attempted to understand the principles of MSDA and analyze its effects. 
Two works~\citep{Mixupcalib1,Mixupcalib2} have found that Mixup helps to calibrate convolutional neural networks (CNNs) and makes them less over-confident. 
Additionally, Wen~\textit{et al.}~\citep{Mixupcalib3} observed that applying both Mixup and model ensemble can be detrimental by making CNNs under-confident. 
Two studies~\citep{Mixuprobust,Mixupdatadepend} provided theoretical analysis on how Mixup-trained models are robust to adversarial attacks and have improved generalization through data-adaptive regularization. 
Li~\textit{et al.}~\citep{Mixupimbal} confirmed that Mixup improves the learning of under-represented classes in class imbalance scenarios that require capturing small structures in semantic segmentation. 
Park~\textit{et al.}~\citep{Mixupcutmix} provided a theoretical analysis on the differences between Mixup and CutMix~\citep{cutmix}.

However, as evidenced in \cref{fig:overview}, the impact of MSDA is not equitably distributed among all classes.
While MSDA improves overall accuracy, it improves performance for some classes while decreasing performance for others. 
The problem of identifying and addressing class dependency or bias is becoming increasingly important for fairness in artificial intelligence and diagnosing rare diseases in clinical applications. 
Most studies on class dependency assume that bias is caused by flaws in the dataset, such as small data and class imbalance. 
However, there are few studies that examine class dependency that is derived from network structures or learning techniques such as data augmentation. 
Recently, Balestriero~\textit{et al.}~\citep{Balestriero2022} reported that conventional data augmentation methods such as image cropping and color jittering could introduce class dependency and also affect after transfer learning.

To reveal the class dependency of MSDA, we propose two metrics for evaluating the class dependency of the data augmentation effect, the number of degraded classes and the average recall change of degraded classes.
Our observations on the class dependency of MSDA lead us to formulate a novel task of reducing class dependency in MSDA. 
In this task, our aim is to minimize performance degradation in the degraded classes while not compromising overall accuracy.
We suggest that the simple technique of blending MSDA and non-MSDA data, termed DropMix, can reduce the class dependency of MSDA effects.
We conduct experiments on two datasets, CIFAR-100 and ImageNet, with three MSDA methods, Mixup, CutMix, and PuzzleMix, to validate the class dependency of MSDA and the effectiveness of DropMix. 
Experiments confirm that the DropMix method not only reduces class dependency but also improves overall average accuracy.

\noindent
The key contributions of this paper are as follows:
\begin{itemize}
 \item We define evaluation metrics to quantify the class dependency of MSDA, the number of degraded classes, and the average degradation performance. With these metrics, we introduce the novel task of reducing class dependency in MSDA.
 \item We investigate the class dependency of MSDA by applying three MSDA methods, Mixup, CutMix, and PuzzleMix to CIFAR-100 and ImageNet classification tasks.
 \item We propose the DropMix method as a straightforward yet effective technique for reducing class dependency and performance degradation in MSDA.
 \end{itemize}

\section{Related Work}
\label{sec:rel}

\subsection{Mixed Sample Data Augmentation (MSDA)}
Mixed sample data augmentation (MSDA) is a technique used to improve the performance of machine learning models by generating new training samples by combining multiple existing samples. 
This is achieved by blending two or more training samples and creating a new sample incorporating features from each of the original samples. 
There are various methods proposed for mixing samples, such as (1) linearly combining two images, e.g. Mixup~\citep{mixup}, (2) replacing a portion of an image with another image, e.g. CutMix~\citep{cutmix} and ResizeMix~\citep{resizemix}, and (3) mixing salient parts of two images, e.g. SaliencyMix~\citep{saliencymix} and PuzzleMix~\citep{puzzlemix}. 
By increasing the diversity of the training data, the model can become more robust to variations in the input data, resulting in improved performance on unseen data. 
Due to these benefits, MSDA is actively applied in deep learning for a variety of applications such as video processing~\citep{msdavideo1,msdavideo2,msdavideo3,msdavideo4}, natural language processing~\citep{msdanlp1,msdanlp3,msdanlp4,msdanlp5,msdanlp6}, and image recognition tasks~\citep{msdaseg1,msdaseg2,msdaseg3,msdaseg4,msdaseg5}.
In this paper, we investigated the class dependency in MSDA by examining three representative methods, Mixup, CutMix, and PuzzleMix, and verified that the proposed DropMix effectively mitigated the class dependency in these MSDA methods.

\subsection{Class Dependency in Deep Learning}
Class dependency or bias in deep learning refers to the phenomenon where a model performs well on certain classes of data but poorly on others. 
This can happen when the model is trained on a dataset that does not accurately represent the population it will be used on, or when the dataset contains inherent biases towards certain classes~\citep{biasmethod0}.
This can result in not only performance issues, such as poor diagnosis of rare diseases, but also social fairness issues such as racial and gender bias. 
Therefore, various methods for analyzing dataset bias~\citep{databias1,databias2,databias3,databias4,databias5} and mitigating class dependency~\citep{biasmethod1,biasmethod2,biasmethod3,biasmethod4,biasmethod5} are being actively researched.

Recent studies have shifted focus from class dependency caused by class imbalance in datasets to class dependency caused by learning techniques such as data augmentation. 
Balestriero~\textit{et al.} found that traditional DA methods, like crop and color jittering, can introduce class dependency through experiments on ImageNet~\citep{Balestriero2022}. 
Building on this research, we examined class dependency in MSDA and proposed a task and method of finding a data augmentation technique that minimizes class bias.

\section{Effect of MSDA is Class Dependent}
\label{sec:classdep}

Motivated by the observations made in \cref{fig:overview}, we first verify whether the effect of MSDA is class-dependent through experiments. 
To demonstrate this, we select Mixup~\citep{mixup}, CutMix~\citep{cutmix}, and PuzzleMix~\citep{puzzlemix} as representative MSDA variants for experimental validation.

\subsection{Experiments}
\textbf{Datasets:} 
We test the Mixup, CutMix and PuzzleMix methods on the CIFAR-100 datasets~\citep{cifar}.
Additionally, the Mixup and CutMix methods are evaluated on the ImageNet-1k dataset~\citep{imagenet}. 
The CIFAR-100 datasets consist of 50,000 training images and 10,000 validation images across 100 classes.
In CIFAR-100, each class has 100 validation images. 
The ImageNet-1k dataset consists of 1.28 million training images and 50,000 validation images across 1,000 classes. 
In ImageNet-1k, each class has 50 validation images.

\noindent
\textbf{Training details:}
For the CIFAR-100 datasets, we trained WideResNet28-2 (WRN28-2)~\citep{wrn}, PreActResNet-18 (PreActRN18)~\citep{preact}, PreActResNet-34 (PreActRN34)~\citep{preact} and PreActResNet-50 (PreActRN50)~\citep{preact} for 300 epochs.
We used Stochastic Gradient Descent with an initial learning rate of 0.2, which decays by a factor of 0.1 at 100 and 200 epochs.
We also set the momentum to 0.9 and the weight decay to 0.0005.
The Mixup and CutMix parameter $\alpha$ is set to 1.0.
For PuzzleMix, we used parameter ($\alpha$, $\beta$, $\gamma$, $\eta$,  $\xi$) = (1.0, 1.2, 0.5, 0.2, 0.8).

For Mixup and CutMix on ImageNet-1k, we trained ResNet-50~\citep{resnet} for 300 epochs. 
Following the training protocol delineated in \citep{cutmix}, we regulated the learning rate over 300 epochs, starting with an initial rate of 0.1 and implementing a decay factor of 0.1 at epochs 75, 150, and 225. We set the batch size to 256.
The Mixup and CutMix parameter $\alpha$ is set to 1.0. 

In addition, we trained vanilla models without the application of MSDA, utilizing identical parameters for comparison.
We trained at least five models with different random seeds and obtained results and metrics by averaging them.

\begin{figure*}
 \centering
 \begin{subfigure}{\linewidth}
 \centering
 \includegraphics[width=\linewidth]{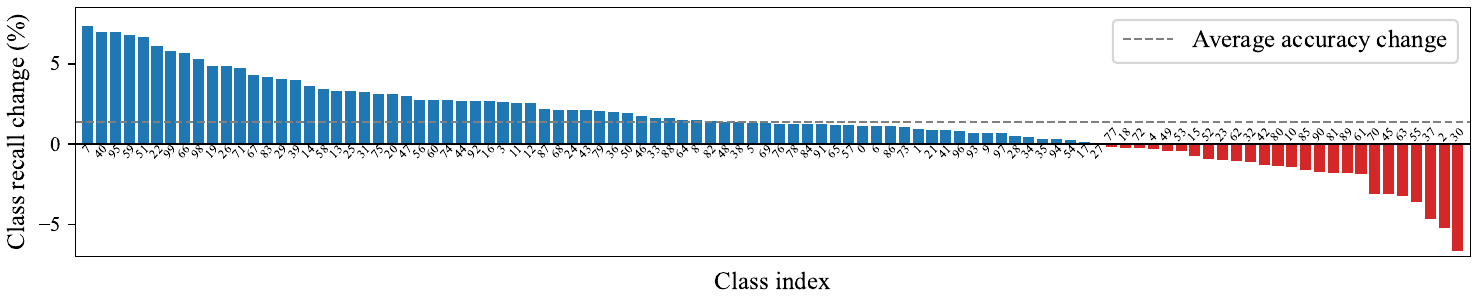}
 \caption{Mixup}
 \label{fig:crc_Mixup_cifar100_Mixup}
 \end{subfigure}
 \hfill
 \begin{subfigure}{\linewidth}
 \centering
 \includegraphics[width=\linewidth]{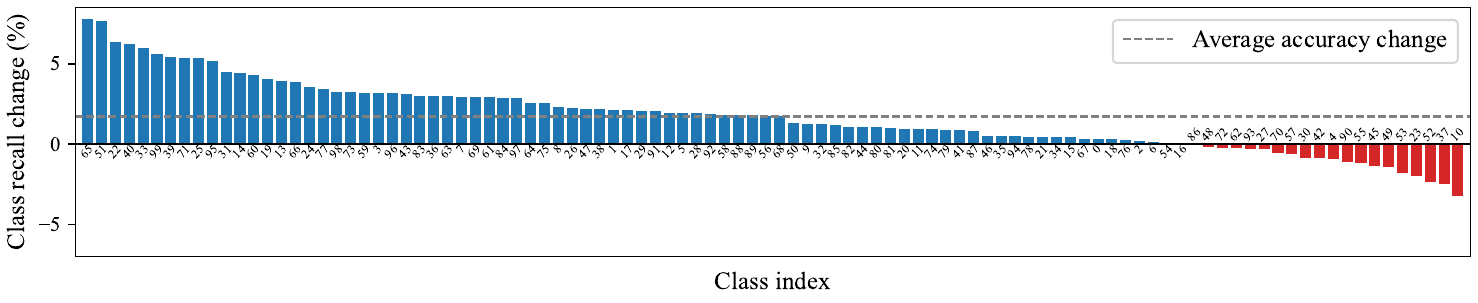}
 \caption{Mixup with DropMix}
 \label{fig:crc_Mixup_cifar100_drMixup}
 \end{subfigure}
 \caption{
 Comparison of class recall changes $\Delta R(m)$ between (a) Mixup and (b) Mixup with DropMix on the CIFAR-100 dataset.
 \textcolor{blue}{Blue} indicates \textcolor{blue}{\textit{improved class}} in which the class recall is improved compared to the vanilla model, while \textcolor{red}{red} indicates \textcolor{red}{\textit{degraded class}} in which the class recall is decreased compared to the vanilla model.
 Classes are arranged in descending order of recall change.
 }
 \label{fig:crc_Mixup_cifar100}
\end{figure*}

\subsection{Evaluation metrics}
We propose some metrics to evaluate class dependency of the classification results of the trained model. 
We first use a \textit{class recall} $R(m)$ as a performance measure at the class level where $m=1,...,M$ is a class index. We now can define a \textit{class recall change} $\Delta R_{MSDA}(m)$:

\begin{equation}
 \Delta R_{MSDA}(m) = R_{MSDA}(m) - R_{vanilla}(m).
 \label{eq:1}
\end{equation}

Using class recall change, we can define sets of classes, \textit{degraded classes} $S_{DC}$:

\begin{equation}
 m\in S_{DC} ~~~~ \text{if} ~~~~ \Delta R_{MSDA}(m) <0.
\end{equation}

With class recall change and degraded class sets, we can define class dependency metrics. 
First, the \textit{number of degraded classes} $N_{DC}$ can be defined as:

\begin{equation}
 N_{DC} = |S_{DC}|.
\end{equation}

Second, the \textit{average recall change of degraded classes} $\overline{\Delta R_{DC}}$ can be defined as:

\begin{equation}
 \overline{\Delta R_{DC}} = \frac{1}{N_{DC}} \sum_{m \in S_{DC}}^{} \Delta R_{MSDA}(m).
\end{equation}

It can be inferred that the effect of MSDA becomes more class-dependent as the number of degraded classes, $N_{DC}$, increases, and the average recall change of degraded classes, $\overline{\Delta R_{DC}}$, decreases. For comparative analysis purpose, metrics for improved classes can also be defined in a similar manner.
We evaluate and analyze the results of experiments using these metrics.
To eliminate randomness in training, the class recall $R(m)$ used to compute the metrics was determined by averaging the $R(m)$ over models trained with different seeds.

\subsection{Results: MSDA Creates Class Dependency}
\Cref{fig:crc_Mixup_cifar100_Mixup} shows the changes in class recall of CIFAR-100 classification model before and after applying Mixup on PreActRN50.
The average accuracy of the model demonstrates a 1.39\%p improvement compared to the vanilla model. 
However, when the performance is evaluated on a class-by-class basis, it is observed that the recall improves for 76 classes, while declining for 24 classes.
Particularly, the ``Beetle'' class has the highest recall improvement of 7.33\%p, while the ``Dolphin'' class has the highest recall decline of 6.67\%p.
This illustrates that Mixup augmentation is accompanied by severe class dependency; not all classes benefit from Mixup during training.

Our investigation on the result suggest that class dependency cannot be explained by simple indicators such as class characteristics or vanilla model performance.
Both the improved and degraded classes contain a diverse range of classes.
Additionally, there is a wide distribution of vanilla model recall in both improved and degraded classes, indicating that the recall range in the vanilla model does not appear to influence class dependency trends.
Visualized investigation recall and class distribution is provided in the supplementary material.
We will examine the intrinsic properties of these difference phenomena by classes later in \cref{sec:discussions}.

The quantitative results for the conditions we experimented with are presented in \cref{tab:cifar,tab:imagenet}. In \cref{tab:cifar}, it can be seen that when various networks are trained with CIFAR-100, the application of various MSDAs, as widely known, yields overall performance improvements, but some classes suffer losses. In some cases, over 30\% of the total classes see a performance decline by MSDA. This trend is also observed in the large-scale dataset represented in \cref{tab:imagenet}. Based on the results from applying Mixup and CutMix to ImageNet and CIFAR-100, we can anticipate that PuzzleMix would exhibit similar dependencies in ImageNet.

\section{DropMix: Dropping Random Samples from MSDA Mitigates Class Dependency}

The observations on class dependency of MSDA can lead to finding an MSDA that is less class dependent, rather than simply improving the overall performance.
We formulate this problem as a task of minimizing the number of degraded classes $N_{DC}$ and amount of their average recall change $\overline{\Delta R_{DC}}$ while maintaining average performance gain.
To achieve this, we proposed DropMix, a simple but efficient method to mitigate the class dependency of MSDA.

\subsection{Methods}

Acknowledging the adverse effects linked exclusively to Mixed MSDA, we argue that entirely forgoing MSDA would disregard its substantial benefits, including enhanced accuracy, generalization, and reduced over-confidence. Therefore, we advocate for a selective application of MSDA, utilizing a DropMix rate for determining which samples to exclude. This strategy involves training excluded samples with their original images and labels, sidestepping MSDA techniques. DropMix randomly selects samples for exclusion without pre-assumed knowledge of class characteristics within datasets or networks. This randomness, aligned with the DropMix rate, ensures a balanced mix of MSDA and non-MSDA samples, simplifying implementation while preserving statistical integrity.

Our approach necessitates no preliminary information regarding the specific MSDA, dataset, and network setup. It allows for the reduction of class dependencies without prior knowledge of which classes may be disadvantaged by the direct application of MSDA. Additionaly, it demands no extra computational resources, rather liberating a portion of MSDA processing in line with the DropMix rate. The detailed, step-by-step procedure for generating samples using MSDA with DropMix is thoroughly delineated in \cref{alg:dropmix}.

\begin{algorithm}
\caption{DropMix}
\label{alg:dropmix}
\begin{algorithmic} 
\STATE \textbf{Input}: DropMix rate $r$, minibatch $X$, $Y$ sampled from data loader
\STATE Generate uniform random number $Rand \in [0,1]$
\IF {$r < Rand$}
\STATE $X, Y \gets MSDA(X, Y)$
\ENDIF
\STATE \textbf{return} $X$, $Y$
\end{algorithmic}
\end{algorithm}

\begin{table*}[b!]
\centering
 \caption{
  Results of the CIFAR-100 experiments. A comparison of the Mixup, CutMix, and PuzzleMix methods was conducted both with and without DropMix across various networks. The variable $r$ denotes the DropMix rate. To evaluate class dependency, we employed proposed metrics such as the \textit{number of degraded classes} $N_{DC}$ and the \textit{average recall change of degraded classes} $\overline{\Delta R_{DC}}$. The class dependency metrics were derived from averaging the class recalls from at least five distinct models, each trained with unique random seeds. The numbers within the parentheses for accuracy indicate the standard deviation among models trained with different random seeds, whereas the numbers within the parentheses for $\overline{\Delta R_{DC}}$ represent the standard deviation across degraded classes. Aside from a few exceptions in accuracy and the number of degraded classes in PuzzleMix (highlighted in \textit{italics}), DropMix effectively enhanced various metrics when compared to the MSDA method devoid of DropMix.
 }
\resizebox{\textwidth}{!}{
\begin{tabular}{cl|ccc|cccc}
\toprule
           &           & \multicolumn{3}{c|}{without DropMix}          & \multicolumn{4}{c}{with DropMix}         \\
Network    & Method    & Acc $\uparrow$        & $N_{DC} \downarrow$   & $\overline{\Delta R_{DC}} \uparrow$ & $r$ & Acc $\uparrow$  & $N_{DC} \downarrow$ & $\overline{\Delta R_{DC}} \uparrow$ \\
\midrule
           & Vanilla   & 74.38(0.17)          & -          & -               & -   & -     & -        & -               \\
           & Mixup     & 75.22(0.15)          & 32         & -3.98(5.76)     & 0.4 & \textbf{76.01}(0.33) & \textbf{21}       & \textbf{-1.58}(1.00)     \\
WRN28-2    & CutMix    & 75.44(0.10)          & 21         & -7.38(10.18)    & 0.5 & \textbf{76.72}(0.27) & \textbf{12}       & \textbf{-2.55}(2.69)     \\
           & PuzzleMix & 76.86(0.11)          & 11         & -4.67(5.34)     & 0.2 & \textbf{77.24}(0.43) & \textbf{9}        & \textbf{-2.29}(3.04)     \\
           \midrule
           & Vanilla   & 76.92(0.22)          & -          & -               & -   & -     & -        & -               \\
           & Mixup     & 78.24(0.10)          & 28         & -1.66(1.60)     & 0.3 & \textbf{78.33}(0.15) & \textbf{24}       & \textbf{-1.54}(1.00)     \\
PreActRN18 & CutMix    & 79.44(0.20)          & 17         & -1.52(1.46)     & 0.3 & \textbf{79.59}(0.24) & \textbf{12}       & \textbf{-0.85}(0.68)     \\
           & PuzzleMix & \textbf{80.12}(0.15) & 9          & \textbf{-1.13}(0.87)     & \textit{0.1} & 79.74(0.25) & \textbf{8}        & -1.48(1.03)     \\ 
           \midrule
           & Vanilla   & 78.06(0.20)          & -          & -               & -   & -     & -        & -               \\
           & Mixup     & 79.54(0.26)          & 29         & -2.17(1.52)     & 0.4 & \textbf{79.67}(0.34) & \textbf{21}       & \textbf{-1.36}(0.99)     \\
PreActRN34 & CutMix    & 80.82(0.20)          & 12         & -1.18(1.35)     & 0.4 & \textbf{81.18}(0.24) & \textbf{6}        & \textbf{-0.83}(0.45)     \\
           & PuzzleMix & 81.56(0.24)          & \textbf{5}            & -1.28(0.75)     & \textit{0.1} & \textbf{81.70}(0.22) & 6        & \textbf{-1.03}(0.63)     \\ 
           \midrule
           & Vanilla   & 78.89(0.12)          & -          & -               & -   & -     & -        & -               \\
           & Mixup     & 80.38(0.24)          & 24         & -2.22(1.54)     & 0.4 & \textbf{80.71}(0.13) & \textbf{18}       & \textbf{-1.52}(0.97)     \\
PreActRN50 & CutMix    & 81.91(0.29)          & 13         & -0.75(0.70)     & 0.4 & \textbf{82.32}(0.14) & \textbf{6}        & \textbf{-0.70}(0.56)     \\
           & PuzzleMix & \textbf{82.92}(0.14) & 6          & -1.73(1.19)     & \textit{0.2} & 82.72(0.11) & \textbf{4}        & \textbf{-1.00}(0.79)     \\
\bottomrule
\end{tabular}
}
\label{tab:cifar}
\end{table*}

\subsection{Experimental Results}
We conducted experiments as outlined in the previous section, utilizing an identical blend of MSDAs, networks, and datasets to investigate class dependency and the effects of DropMix. Our results demonstrate that class dependency occurs in various scenarios and that the proposed DropMix method effectively mitigates it.

\textbf{CIFAR-100 results: }
The results of our experiments on the CIFAR-100 datasets, utilizing the Mixup, CutMix, and PuzzleMix methods, without and with DropMix are presented in \cref{tab:cifar}. 
In all the networks we tested with Mixup and CutMix, the models with DropMix demonstrated superior accuracy, fewer degraded classes, and better recall degradation compared to those without DropMix. The introduction of original samples in a random manner during the learning process enhance the label-preserving tendency of the classes, thereby reducing model dependency. In such cases, mitigating the degradation caused by bias has an overall positive impact on performance. However, in the PuzzleMix experiment, we did not see improvements in the accuracy for PreActRN18 and PreActRN50, nor in the average change in class recall of degraded classes for PreActRN18. Firstly, the evaluation metrics for the class dependency of PuzzleMix without DropMix across all networks were observed to be better than those of Mixup and CutMix. Given that PuzzleMix employs the saliency map of samples in a manner that is more label-preserving and less biased, the impact of the label-preserving samples derived from the application of DropMix is less significant. Therefore, we can interpret that the effect of DropMix on PuzzleMix is diminished due to the lower impact of label-preserving samples from DropMix.

\begin{table*}[b!]
\centering
 \caption{
   Results of the ImageNet experiments. A comparison of the Mixup and CutMix methods was conducted both with and without DropMix on ResNet50. The style shown in \cref{tab:cifar} is used for formatting and notation. The class dependency metrics were derived from averaging the class recalls from five distinct models, each trained with unique random seeds.
 }
 \resizebox{\textwidth}{!}{
\begin{tabular}{cl|ccc|cccc}
\toprule
         &         & \multicolumn{3}{c|}{without DropMix} & \multicolumn{4}{c}{with DropMix}         \\
Network  & Method  & Acc $\uparrow$   & $N_{DC} \downarrow$  & $\overline{\Delta R_{DC}} \uparrow$ & $r$ & Acc $\uparrow$  & $N_{DC} \downarrow$ & $\overline{\Delta R_{DC}} \uparrow$ \\
\midrule
         & Vanilla & 76.89(0.14)  & -         & -               & -   & -     & -        & -               \\
ResNet50 & Mixup   & 77.75(0.11)  & 332       & -2.19(1.97)     & 0.1 & \textbf{77.91}(0.12) & \textbf{316}      & \textbf{-2.13}(1.90)    \\
         & CutMix  & 78.52(0.04)  & 225       & -2.01(1.93)     & 0.1 & \textbf{78.70}(0.09) & \textbf{187}      & \textbf{-1.98}(1.71)    \\
\bottomrule
\end{tabular}
}
\label{tab:imagenet}
\end{table*}

\textbf{ImageNet results: }
The results of our experiments on the Mixup and CutMix based models, using the ImageNet-1k dataset, are presented in \cref{tab:imagenet}. The efficacy of DropMix is pronounced even in large-scale datasets. Compared to vanilla model, Mixup and CutMix experienced recall degradation in 332 and 225 classes, respectively. By applying a DropMix rate of 0.1, this was reduced to 316 and 187 classes, and there was also an improvement in the average recall changes of degraded classes. Moreover, in both MSDA methods, these effects were accompanied by an increase in average accuracy by 0.16\%p and 0.18\%p. Additionally, DropMix was effective for both Mixup and CutMix in terms of the average change in class recall of degraded classes.

\begin{figure*}[t]
 \centering
 \begin{subfigure}{0.31\linewidth}
 \centering
 \includegraphics[width=\linewidth]{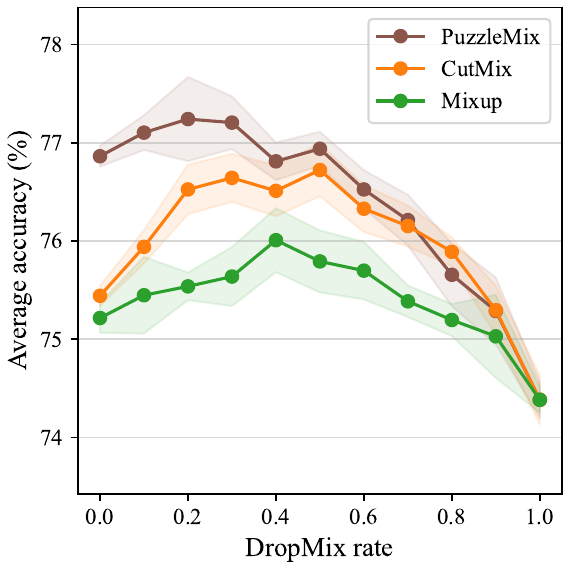}
 \caption{Average accuracy}
 \label{fig:avg}
 \end{subfigure}
 \quad
 \begin{subfigure}{0.31\linewidth}
 \centering
 \includegraphics[width=\linewidth]{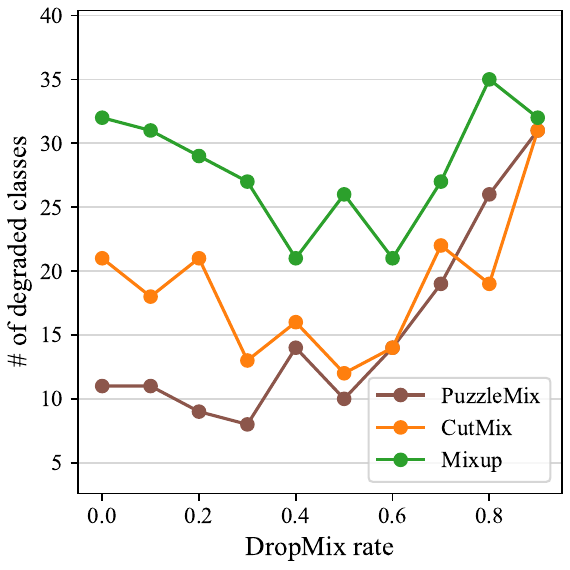}
 \caption{$N_{DC}$}
 \label{fig:deg}
 \end{subfigure}
 \hfill
 \begin{subfigure}{0.31\linewidth}
 \centering
 \includegraphics[width=\linewidth]{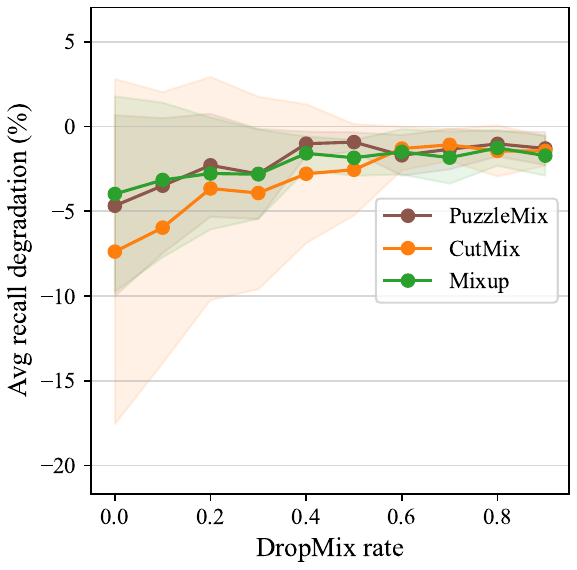}
 \caption{$\overline{\Delta R_{DC}}$}
 \label{fig:avg recall deg}
 \end{subfigure}
 \caption{
 The performances of Mixup (represented in green), CutMix (represented in orange), and PuzzleMix (represented in brown) with DropMix, respectively, are analyzed in terms of class-dependent metrics as a function of the DropMix rate. This analysis was conducted on the WideResNet28-2 trained with CIFAR-100 dataset. In (a), the average accuracy of the DropMix method surpasses that of the respective MSDA methods (where the DropMix rate is 0) for relatively small DropMix rates. In (b), the number of degraded classes $N_{DC}$ is reduced for a small DropMix rate across all MSDA methods. In (c), the average recall change of degraded classes $\overline{\Delta R_{DC}}$ is mitigated across the entire range of DropMix. The shaded region in (a) illustrates the standard deviation among five distinct models, while the shaded region in (b) depicts the standard deviation across degraded classes.
 }
 \label{fig:ablation_cifar100}
\end{figure*}

\textbf{Ablation study:}
\cref{fig:ablation_cifar100} graphically demonstrates the variation of class dependency metrics with changes in the hyperparameter, DropMix rate, during the training of WRN28-2 on the CIFAR-100 dataset.
In accordance with the established interpretation of the DropMix rate, applying DropMix with a rate of 0 is equivalent to employing MSDA without the DropMix, while a rate of 1 corresponds to vanilla training without the application of MSDA.
As depicted in \cref{fig:avg}, the average accuracy demonstrates a trend of decreasing from MSDA to the vanilla model as the DropMix rate increases. 
However, in the region of small values, it surpasses the average accuracy of MSDA without DropMix. 
This can be attributed to the benefits of dependency reduction in the same region, as depicted in \cref{fig:deg,fig:avg recall deg}.

If the application of DropMix leads to an increase in model accuracy, the DropMix rate at which this value peaks was considered the optimal rate. 
In this study, we selected the DropMix rate that yielded the best average accuracy. 
Under the conditions where Mixup and CutMix are applied, the best average accuracy in almost all scenarios is consistent with the best class dependency mitigation conditions, indicating that class dependency mitigation has a direct positive impact on overall performance.
In cases where there was no increase in average accuracy, we selected the rate that corresponded to a reduction in the number of degraded classes or an improvement in the average recall change of these classes.
Concurrent with the increase in average accuracy under the Mixup and CutMix methods, we observed a decrease in the number of degraded classes and an increase in average recall change, confirming the mitigation of class dependency.

In the case of PuzzleMix, there were instances where, despite the metrics indicating bias mitigation showing improvement, the average accuracy did not necessarily increase. 
The particular DropMix rates corresponding to this observation are highlighted in \textit{italic} in \cref{tab:cifar}.
The results of similar ablation procedures for CIFAR-100 experiments across other networks are provided in the supplementary material.
As previously stated, PuzzleMix acquires the ability to preserve label information through the salience information, but at the cost of additional computing power.
\section{Discussions}
\label{sec:discussions}

\begin{figure*}[t!]
 \centering
  \begin{subfigure}{0.48\linewidth}
     \includegraphics[width=0.9\linewidth]{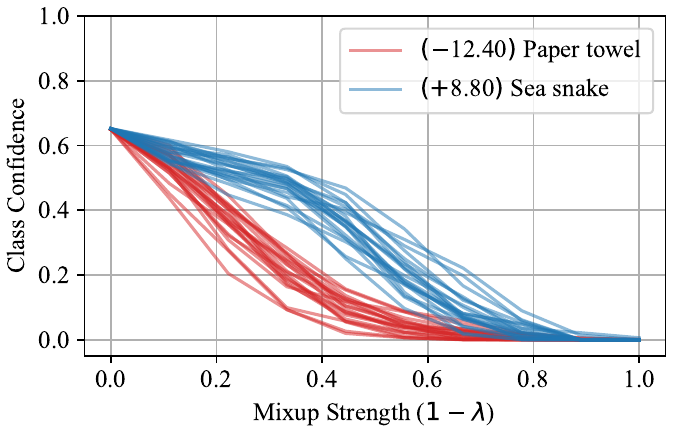}
     \caption{}
 \end{subfigure}
  \begin{subfigure}{0.48\linewidth}
     \includegraphics[width=0.9\linewidth]{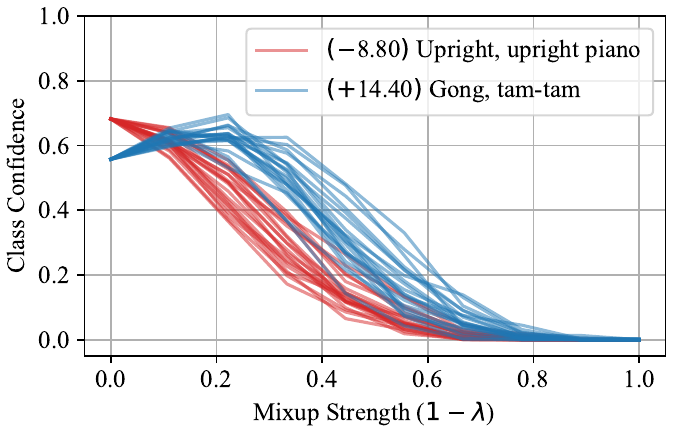}
     \caption{}
 \end{subfigure}
 \caption{
    The impact of Mixup strength on the label information. Each line represents a different class as counterparts in Mixup. The value before the class name indicates the class recall change before and after Mixup, observed in our experiment. As the strength of Mixup increases, the label information for the class is progressively lost. This loss occurs at a noticeably faster rate for the \textcolor{red}{\textit{degraded class}} compared to the \textcolor{blue}{\textit{improved class}}. These findings were obtained from the validation set of ImageNet using the official PyTorch ResNet50 model, which has been pre-trained on the ImageNet dataset.
 }
 \label{fig:mixup_intensity}
\end{figure*}

\subsection{Why does MSDA create class dependency?}
Balestriero~\textit{et al.} has proven that the non label-preserved augmentation can bring inevitable bias to the learned model~\citep{Balestriero2022}.
Given that MSDA methods, such as Mixup, CutMix, and PuzzleMix, exhibit non-label-preserving characteristics during the label smoothing process, the class dependency observed in our experimental results is consistent with Balestriero's theorem.

\textbf{The impact of Mixup strength on the label information: } Does applying MSDA affect label information differently across degraded and improved classes? To explore this, we analyzed label information of Mixup-applied samples using a pre-trained ResNet in \cref{fig:mixup_intensity}. 
Findings from this study reveal that in most cases, classes negatively impacted by Mixup experience a more rapid loss of class label information compared to those that benefit. 
Although label information loss is inevitable with MSDA, the differential rates of loss among various classes can create class dependency.
The observation suggests that the phenomenon of degraded classes is determined more by the intrinsic characteristics of the classes rather than being a random occurrence. 
This underscores the need for further research to understand the reasons behind the varying vulnerability of classes to information loss during MSDA.
Impact analysis for a wider range of combinations will be provided in supplementary material. 

\begin{figure*}[t!]
 \centering
 \begin{subfigure}{0.49\linewidth}
 \centering
     \includegraphics[width=0.7\linewidth]{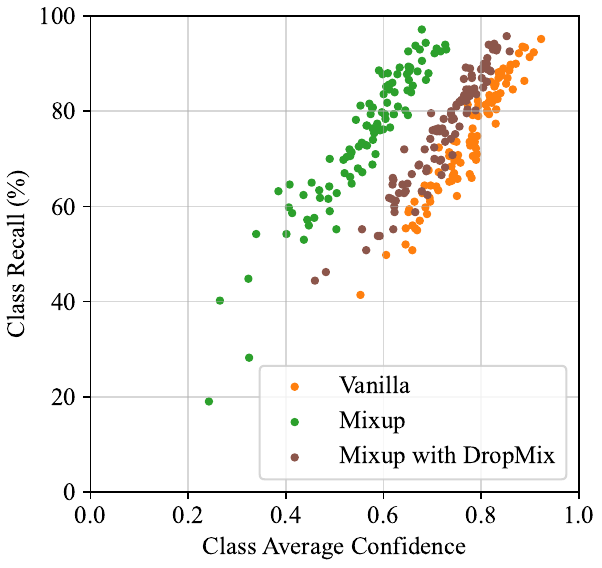}
     \caption{Class confidence v.s. $R$}
     \label{fig:confdence_recall1}
 \end{subfigure}
  \begin{subfigure}{0.49\linewidth}
  \centering
     \includegraphics[width=0.68\linewidth]{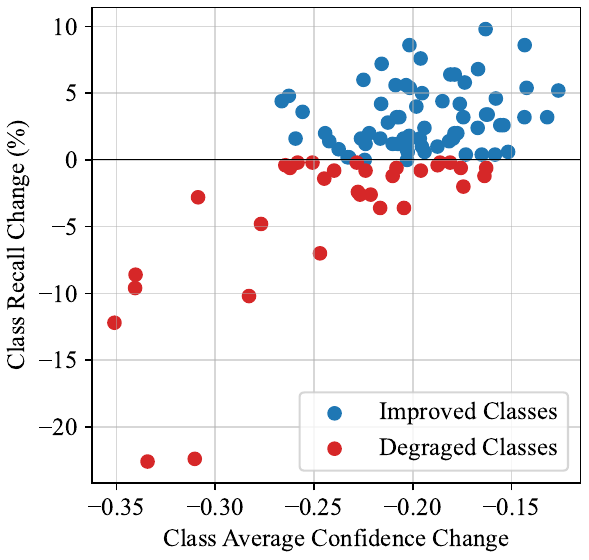}
     \caption{Class confidence change v.s. $\Delta R_{Mixup}$}
     \label{fig:confdence_recall2_change}
 \end{subfigure}
 \caption{
    Correlation between class confidence and recall. Each dot represents an individual class. This analysis explores the correlation between class confidence and recall, utilizing individual class data from WideResNet28-2 models trained on the CIFAR-100 dataset. In (a), Mixup (represented in green) reduces confidence levels in comparison to the vanilla model (represented in orange). However, the integration of DropMix (represented in brown) offsets this confidence shift and helps underperforming classes. In (b), there is a significant correlation in which classes with severe recall and confidence degradation problems cluster in the lower left quadrant. The results of five models, each trained with different random seeds, were averaged to reach this figure.
 }
 \label{fig:confidence_recall}
\end{figure*}
\begin{table}[b!]
\centering
 \caption{
   Comparative analysis of class average confidence change in degraded ($DC$) and improved classes ($IC$) across all CIFAR-100 experiments utilizing identical models in \cref{tab:cifar}. This table encapsulates the differential impact on confidence levels, highlighting a greater decrease in $DC$ as opposed to $IC$
 }
\begin{tabular}{l|cc|cc|cc|cc}
\toprule
           & \multicolumn{2}{c|}{WRN28-2} & \multicolumn{2}{c|}{PreActRN18} & \multicolumn{2}{c|}{PreActRN34} & \multicolumn{2}{c}{PreActRN50} \\
Method     & $IC$         & $DC$        & $IC$          & $DC$        & $IC$          & $DC$        & $IC$          & $DC$        \\
\midrule
Mixup      & -0.195     & {\bf -0.241}    & -0.172      & {\bf -0.191}    & -0.167      & {\bf -0.195}    & -0.185      & {\bf -0.211}    \\
CutMix     & -0.134     & {\bf -0.238}    & -0.071      & {\bf -0.088}    & -0.011      & {\bf -0.026}    & -0.076      & {\bf -0.081}    \\
PuzzleMix  & -0.092     & {\bf -0.160}    & -0.053      & {\bf -0.069}    & -0.026      & {\bf -0.055}    & -0.077      & {\bf -0.103}    \\
\bottomrule
\end{tabular}
\label{tab:confidence}
\end{table}

\begin{figure}[t!]
 \centering
 \includegraphics[width=0.6\linewidth]{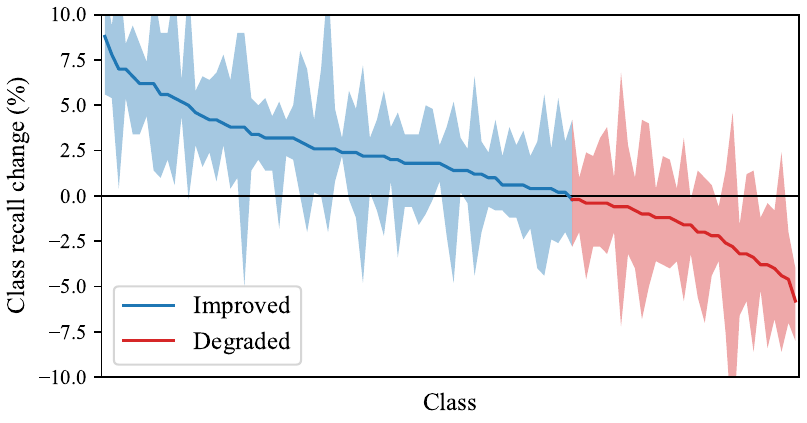}
 \caption{
 The class recall changes of Mixup on the ResNet18 trained with CIFAR-100 dataset, sorted in descending order. The shaded region in illustrates the range of class recall changes obtained from multiple models trained with different random seeds, while the solid lines indicate the average of it. 
 }
 \label{fig:crc2_Mixup_cifar100}
\end{figure}

\textbf{Class confidence-recall correlation: }Based on these findings, we examined the confidence alongside the recall of each class with WRN28-2 on CIFAR-100. MSDA is known to enhance robustness and generalization while mitigating over-confidence in contemporary deep learning models~\citep{mixup_gen}. In \cref{fig:confdence_recall1}, we can observe the average confidence and recall correlation for each class using vanilla, Mixup, and Mixup with DropMix techniques. Overall, Mixup significantly lowered confidence, isolating some classes far from others. \Cref{fig:confdence_recall2_change} reveals from a recall and confidence change perspective that classes with severe recall degradation also suffered substantial confidence loss. Classes that were classified as degraded, but to a lesser extent, were in a similar distribution to most other classes, and these are the classes in the central part of \cref{fig:crc2_Mixup_cifar100}. Returning to \cref{fig:confdence_recall1}, the application of DropMix mitigated the overall loss of both recall and confidence, and brought the classes that suffered severe losses closer to the main distribution.

This observation is significant not only for the individual classes that suffered drastic performance degradation, but also when evaluated as an average of the groups designated as degraded and improved classes. The extension of this analytical approach to all CIFAR-100 experiments is systematically reported in \cref{tab:confidence}. A consistent pattern emerges across the various experimental conditions, where the magnitude of the decline in confidence in the degraded class exceeds that observed in the improved class, consistent with the individual class analysis.

\textbf{Class dependency from randomness: }However, expecting ideal learning conditions in real-world scenarios is impractical, and the observed class dependency might stem from other factors, such as varying initializations or random number generation processes, as exemplified in other contexts.
To ascertain that the observed class dependency is not an artifact of a specific randomness of computation, we conducted experiments with five different random seeds, thereby ensuring that the class dependency persists across varying levels of randomness during the training process.
\Cref{fig:crc2_Mixup_cifar100} shows the classes that went through gradation in class recall change in the ResNet18~\citep{resnet} trained on the CIFAR-100 dataset using Mixup.
This graph illustrates the distribution and trend of class recall changes for Mixup models trained with different random seeds, relative to the average class recall of the vanilla models.
Some classes are distributed in narrow regions with different individual model results, while others are distributed in wide regions. 
This implies that MSDA is not the only cause of the class-dependency phenomenon we observed and is difficult to disentangle. 
At the same time, the solid line trend in \cref{fig:crc2_Mixup_cifar100}, representing the average change in class recall, reveals a clear class-dependent effect of MSDA.

\subsection{DropMix}
The key element of DropMix is to partially expose hard-label samples to reduce class dependency during MSDA-assisted learning. 
Similar methods of varying the frequency of MSDA have been proposed in the past; 
Inoue~\textit{et al.} proposed a method of alternating batches with and without Mixup-like sample pairing~\citep{Inoue2018}. 
In the official CutMix code~\footnote{https://github.com/clovaai/CutMix-PyTorch}, a variable called CutMix probability regulates the frequency of CutMix in a batch. 
Touvron~\textit{et al.} used a Mixup and CutMix probability of 0.8 and 1, respectively, to train data-efficient transformers~\citep{Touvron2021}. 
However, they treated the Mixup or CutMix probability as a hyperparameter for fine-tuning and did not examine its significance or impact.a
Our contribution and novelty lie in proposing a method for adjusting MSDA frequency to reduce class dependency and show its effectiveness from a perspective of class dependency.

\begin{figure*}[t]
 \centering
 \begin{subfigure}{0.31\linewidth}
 \centering
 \includegraphics[width=\linewidth]{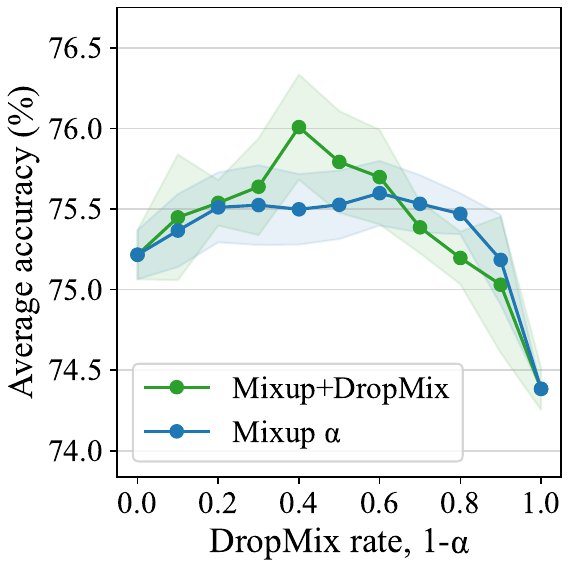}
 \caption{Average accuracy}
 \label{fig:avg_alpha}
 \end{subfigure}
 \quad
 \begin{subfigure}{0.31\linewidth}
 \centering
 \includegraphics[width=\linewidth]{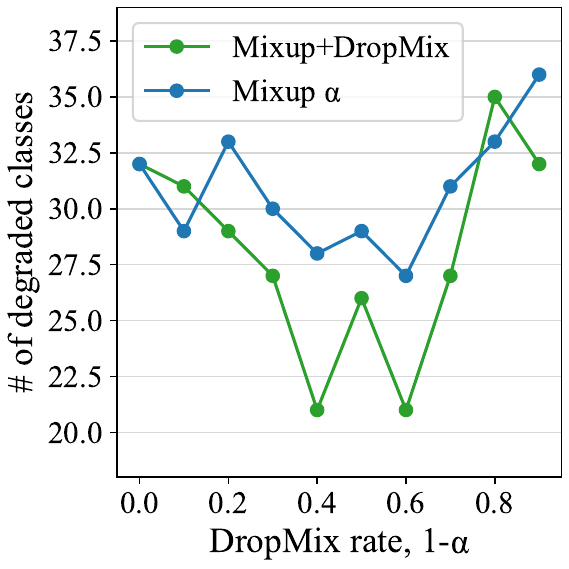}
 \caption{$N_{DC}$}
 \label{fig:deg_alpha}
 \end{subfigure}
 \hfill
 \begin{subfigure}{0.31\linewidth}
 \centering
 \includegraphics[width=\linewidth]{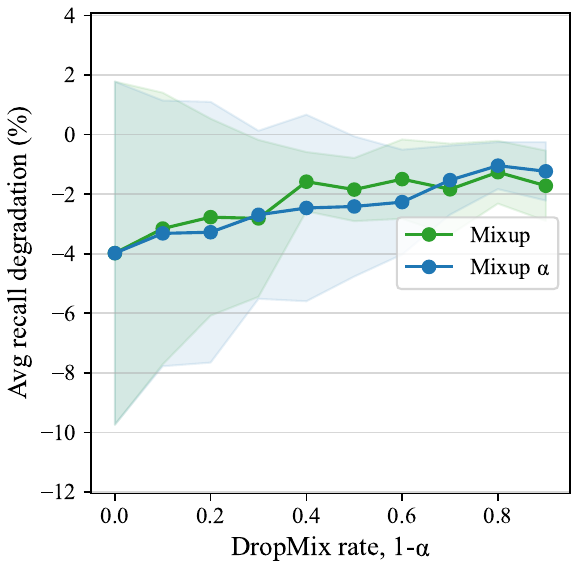}
 \caption{$\overline{\Delta R_{DC}}$}
 \label{fig:recall_alpha}
 \end{subfigure}
 \caption{
 Mixup with DropMix (represented in green), and $\alpha$-controlled (represented in blue). This analysis was conducted on the WideResNet28-2 trained on the CIFAR-100 dataset, following format depicted in \cref{fig:ablation_cifar100}.
 }
 \label{fig:cifar100_alpha}
\end{figure*}

\textbf{How hyperparameter-controlled MSDA differs from DropMix:} In terms of methodology, DropMix incorporates non-mixed samples, whereas $\alpha$-controlled MSDA does not. When considering the distribution of $\lambda$ used in mixing, DropMix exhibits a new type of distribution that includes 0 and 1, rather than a beta distribution with $\alpha$. 
This aspect introduces a variety during training.
To compere their effects, we experimented $\alpha$-controlled Mixup and visualized in \cref{fig:cifar100_alpha} aligning the context of $\alpha$ and DropMix rate by using $1-\alpha$ as the x-axis.
As observed in the \cref{fig:cifar100_alpha}, $\alpha$-controlled exhibits lower overall accuracy gain compared to DropMix and does not show significant improvement in the number of degraded classes.

\subsection{Open Problems}
Our experiments revealed that MSDA methods exhibit class dependency, which was successfully mitigated by the proposed DropMix technique.
However, there were variations in the degree of class dependency and the effectiveness of DropMix among the methods. Mixup demonstrated the highest level of class dependency and showed the greatest improvement with DropMix, followed by CutMix and PuzzleMix. We also discovered a significant correlation between changes in class accuracy and confidence and a difference in label information loss in MSDA. One limitation of our study is the lack of analysis regarding the reasons behind these differences in class dependency and the effects of DropMix across the different MSDA methods. 

We propose two hypotheses to explain this phenomenon: First, MSDA methods with superior overall performance tend to exhibit lower class dependency. Second, local MSDA methods such as CutMix and PuzzleMix, which have a higher probability of preserving label information, have less class dependency compared to the global MSDA method, Mixup. We anticipate that further experiments and analyses will validate our hypotheses and provide insights into the underlying reasons for these observations.

Additionally, future research directions could include: (1) Further investigation on factors that influence the identification of degraded classes, including imaging characteristics, (2) Proposing tailored approaches to minimize class dependency in MSDAs, (3) Extending the study of class dependency in MSDAs to other fields, such as natural language processing.

We commit to releasing the code used in our study to the research community upon the acceptance of our paper.
\section{Conclusion}
\label{sec:conc}

In this paper, we investigated the class dependency of the MSDA effect in image recognition tasks. To examine this, we applied Mixup, CutMix, and PuzzleMix to CIFAR-100 and ImageNet classification datasets, leading to the confirmation that MSDA methods exhibit class dependency. To quantify the degree of class dependency, we introduced two evaluation metrics: the number of degraded classes and the average degradation performance. To address this class dependency in MSDA, we proposed the DropMix technique, and experimental results demonstrated that DropMix effectively reduced class dependency while also enhancing classification efficiency. 
Additionally, we provide a thorough analysis and discussion of the causes of class dependence resulting from MSDA, identifying the classes that are most affected by this dependence.
Our work stands out as the first to propose a quantitative measure for estimating the degree of class dependency in MSDA and a technique to mitigate it.

\subsubsection*{Acknowledgments}
This research was supported by Basic Science Research Program through the National Research Foundation of Korea (NRF) funded by the Ministry of Education (2022R1I1A1A01071970).

\bibliography{MAIN}
\bibliographystyle{iclr2021_conference}

\newpage
\appendix

\begin{table*}[b!]
\centering
 \caption{
  Results of the CIFAR-10 experiments. A comparison of the Mixup and CutMix methods was conducted both with and without DropMix on ResNe t18. The variable $r$ denotes the DropMix rate. To evaluate class dependency, we employed proposed metrics such as the number of degraded classes ($N_{DC}$) and the average change in class recall for degraded classes ($\overline{\Delta R_{DC}}$). All metrics were derived by averaging the outcomes from five distinct models, each trained with unique random seeds.
 }
\begin{tabular}{cc|ccc|cccc}
\toprule
           &           & \multicolumn{3}{c|}{without DropMix}          & \multicolumn{4}{c}{with DropMix}         \\
Network    & Method    & Acc            & $N_{DC}$   & $\overline{\Delta R_{DC}}$ & $r$ & Acc   & $N_{DC}$ & $\overline{\Delta R_{DC}}$ \\ \hline
           & Vanilla   & 95.38          & -          & -               & -   & -     & -        & -               \\
ResNet18   & Mixup     & 96.01          & 1         & -1.40            & 0.1 & 96.13 & 1       & -0.70     \\
           & CutMix    & 96.45          & 0         & -               & 0.2 & 96.56 & 0       & -     \\
\bottomrule
\end{tabular}
\label{tab:cifar10}
\end{table*}

\section{CIFAR-10 Experiment}
\label{sec:add_fig}

In addition to CIFAR-100 and ImageNet datasets, we also conducted a measurement of class dependency of MSDA for CIFAR-10. 
We trained ResNet18 models on the CIFAR-10 dataset for 200 epochs. 
We used Stochastic Gradient Descent with an initial learning rate of 0.1, which decays by a factor of 0.2 at 60, 120, and 160 epochs. 
We also set the momentum to 0.9 and the weight decay to 0.0005. 
The Mixup parameter $\alpha$ is set to 1.0. 
We also trained vanilla models using the same parameters. 
We trained five distinct models in total and obtained results and metrics by averaging them.

\Cref{tab:cifar10} summarized the CIFAR-10 experiment on class dependency for MSDA methods and our DropMix methods.
It was observed that one specific class experienced a decline in the recall. However, through the application of DropMix, this degradation was effectively mitigated, resulting in an improvement from -1.40 to -0.70 and an enhancement in overall average accuracy. Conversely, when CutMix was applied, no noticeable decline in recall at the class level was observed. Nevertheless, DropMix still contributed to an increase in average accuracy. These results are summarized in \cref{tab:cifar10}.

\section{Additional Figures}
\label{sec:add_fig}

This section includes additional figures of class dependency observations and DropMix results discussed in our paper.
These figures, which are variations of the figures presented in the text, demonstrate that the class dependency of MSDA and the effectiveness of the proposed DropMix method have a consistent tendency regardless of the dataset and MSDA method used. 

\begin{figure}[t!]
 \centering
 \begin{subfigure}{0.48\linewidth}
  \includegraphics[width=\linewidth]{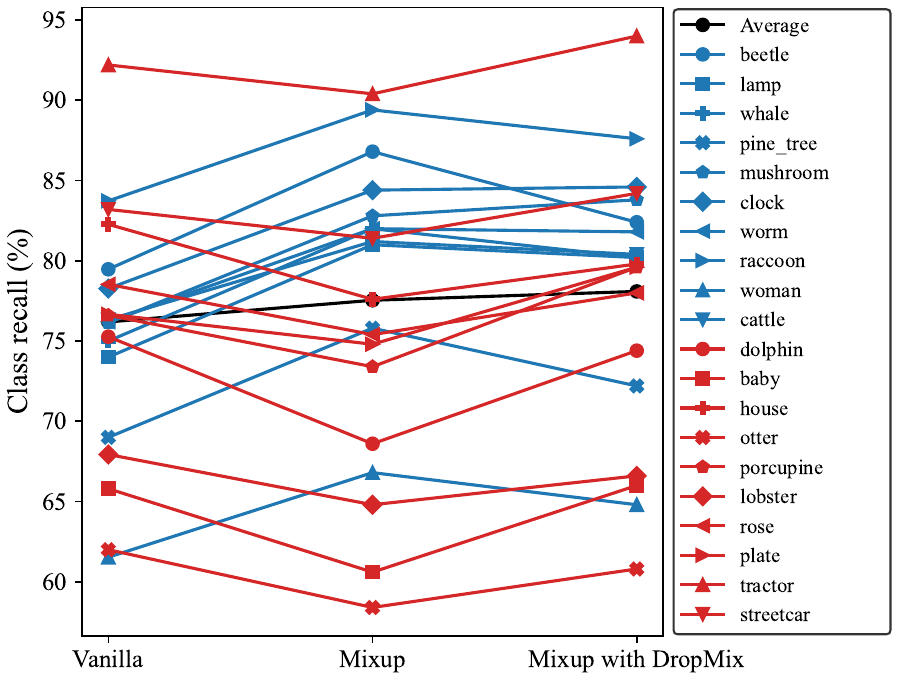}
  \caption{CIFAR-100 with PreActRN50}
 \end{subfigure}
 \begin{subfigure}{0.48\linewidth}
  \includegraphics[width=\linewidth]{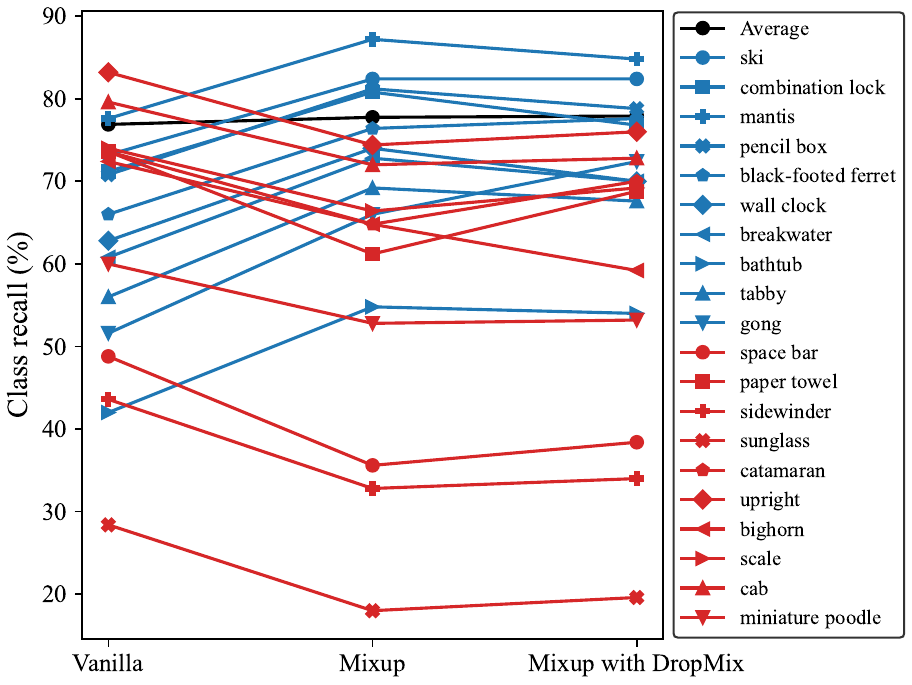}
  \caption{ImageNet with ResNet50}
 \end{subfigure}
 \caption{
 The class recalls of the vanilla model, Mixup, and Mixup with DropMix method were compared on CIFAR-100 and ImagenNet datasets. 
 The top 10 \textcolor{blue}{\textit{improved class}} and the top 10 \textcolor{red}{\textit{degraded class}} were selected. 
 The average accuracy (in black) of each model is also displayed. 
 The results demonstrate that DropMix effectively reduced recall loss in the worst degraded classes while minimally affecting the best-improved classes, leading to an overall increase in average accuracy.
 }
 \label{fig:cr_ex_mixup}
\end{figure}

\begin{figure}[t!]
 \centering
 \begin{subfigure}{0.48\linewidth}
  \includegraphics[width=\linewidth]{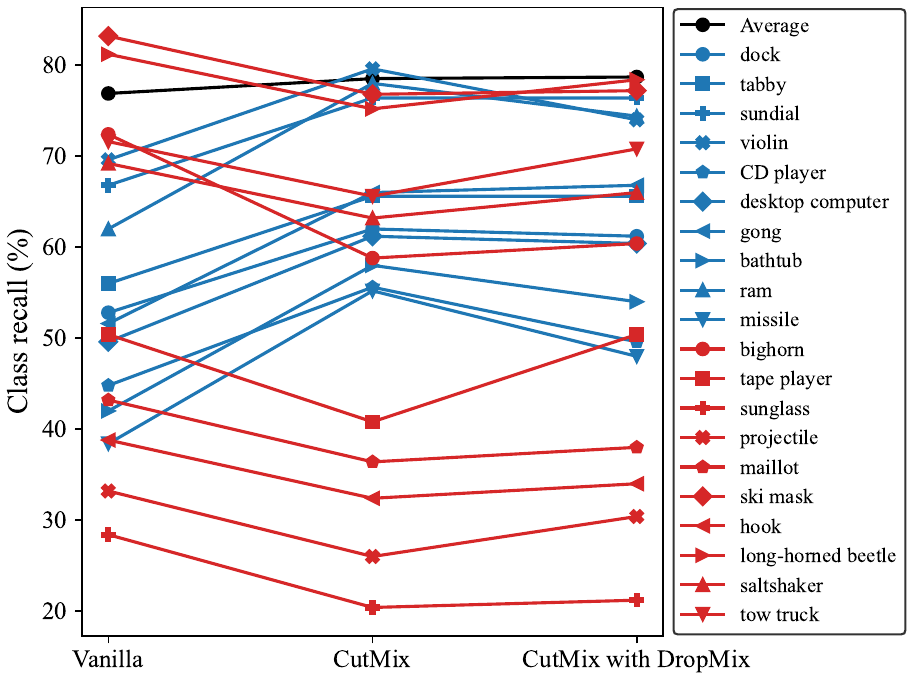}
  \caption{ImageNet}
 \end{subfigure}
 \caption{
  The class recalls of the vanilla model, CutMix method, and CutMix with DropMix method were compared on ImageNet with ResNet50 combination. 
 The top 10 \textcolor{blue}{\textit{improved class}} and the top 10 \textcolor{red}{\textit{degraded class}} were selected. 
 The average accuracy (in black) of each model is also displayed. 
 The results demonstrate that DropMix effectively reduced recall loss in the worst degraded classes while minimally affecting the best improved classes, leading to an overall increase in average accuracy.
 Class index was marked for distinguishing two similar classes. 
 }
 \label{fig:cr_ex_cutmix}
\end{figure}

\begin{figure*}[t!]
 \centering
 \begin{subfigure}{\linewidth}
 \centering
 \includegraphics[width=\linewidth]{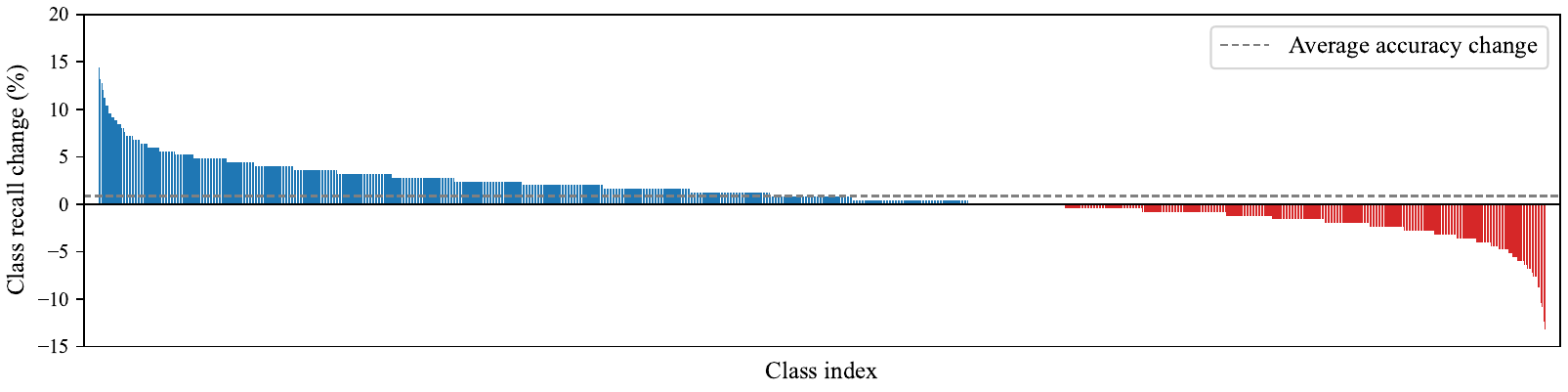}
 \caption{Mixup}
 \label{fig:crc_Mixup_imgnet_Mixup}
 \end{subfigure}
 \hfill
 \begin{subfigure}{\linewidth}
 \centering
 \includegraphics[width=\linewidth]{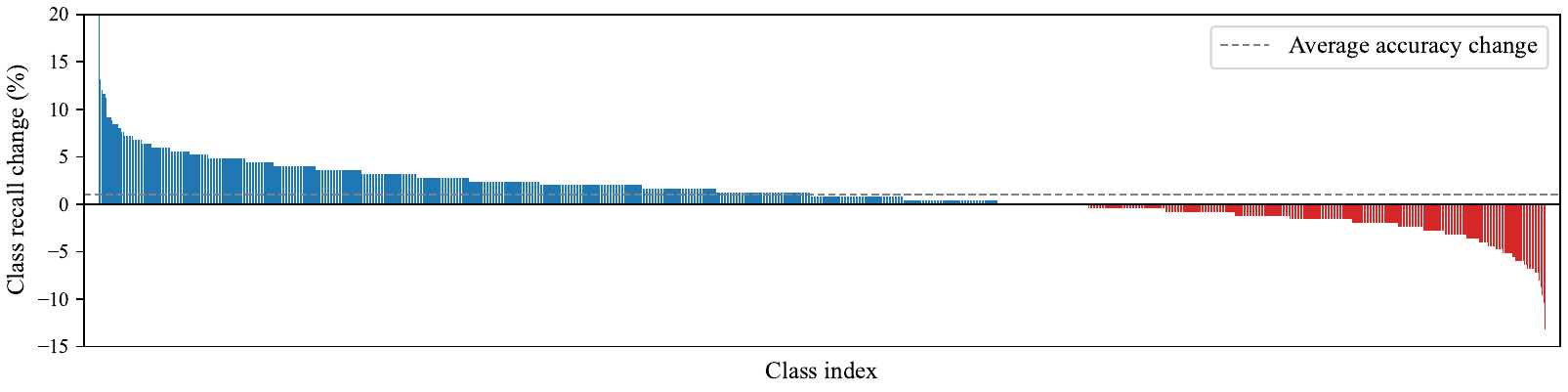}
 \caption{Mixup with DropMix}
 \label{fig:crc_Mixup_imgnet_drMixup}
 \end{subfigure}
 \caption{
Comparison of class recall changes $\Delta R(m)$ between (a) Mixup and (b) Mixup with DropMix on the ImageNet dataset with ResNet50.
 \textcolor{blue}{Blue} indicates \textcolor{blue}{\textit{improved class}} in which the class recall is improved compared to the vanilla model, while \textcolor{red}{red} indicates \textcolor{red}{\textit{degraded class}} in which the class recall is decreased compared to the vanilla model.
 Classes are arranged in descending order of recall change.
 }
 \label{fig:crc_mixup_imagenet}
\end{figure*}

\begin{figure*}[t!]
 \centering
 \begin{subfigure}{\linewidth}
 \centering
 \includegraphics[width=\linewidth]{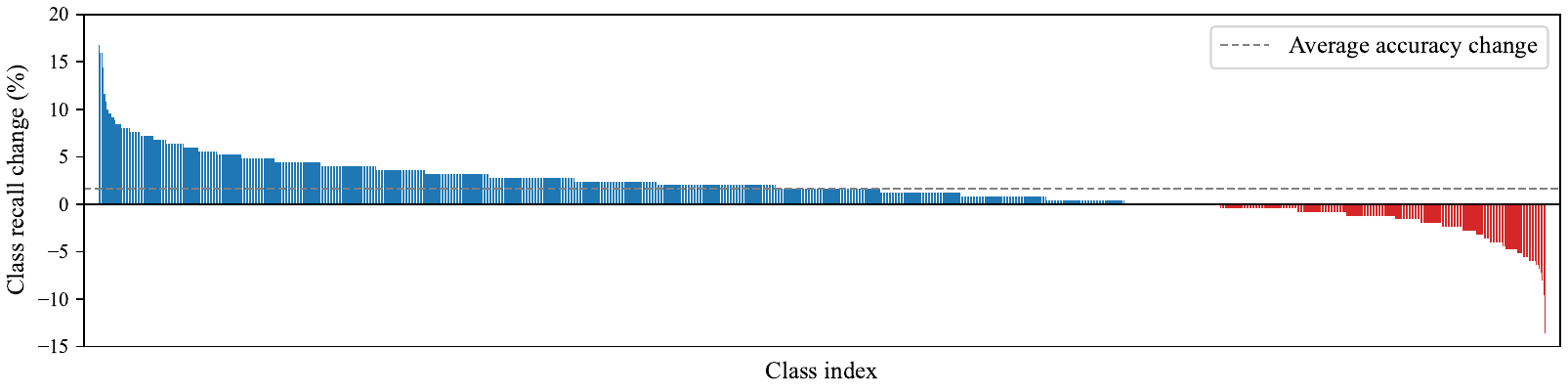}
 \caption{CutMix}
 \label{fig:crc_CutMix_imgnet_CutMix}
 \end{subfigure}
 \hfill
 \begin{subfigure}{\linewidth}
 \centering
 \includegraphics[width=\linewidth]{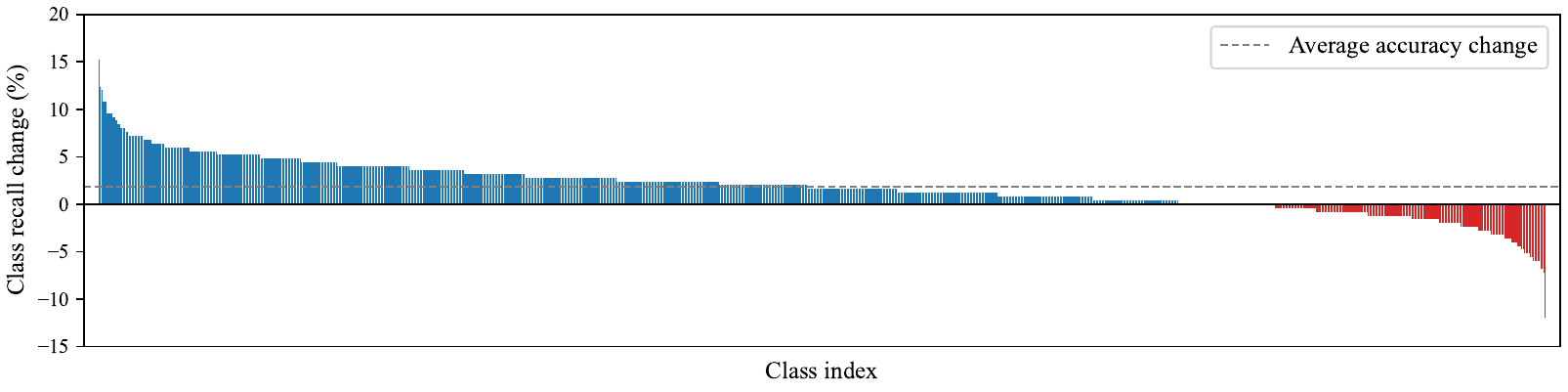}
 \caption{CutMix with DropMix}
 \label{fig:crc_CutMix_imgnet_drCutMix}
 \end{subfigure}
 \caption{
Comparison of class recall changes $\Delta R(m)$ between (a) CutMix and (b) CutMix with DropMix on the ImageNet dataset with ResNet50.
 \textcolor{blue}{Blue} indicates \textcolor{blue}{\textit{improved class}} in which the class recall is improved compared to the vanilla model, while \textcolor{red}{red} indicates \textcolor{red}{\textit{degraded class}} in which the class recall is decreased compared to the vanilla model.
 Classes are arranged in descending order of recall change.
 }
 \label{fig:crc_cutmix_imagenet}
\end{figure*}

\begin{figure*}
 \centering
 \begin{subfigure}{0.29\linewidth}
 \includegraphics[width=\linewidth]{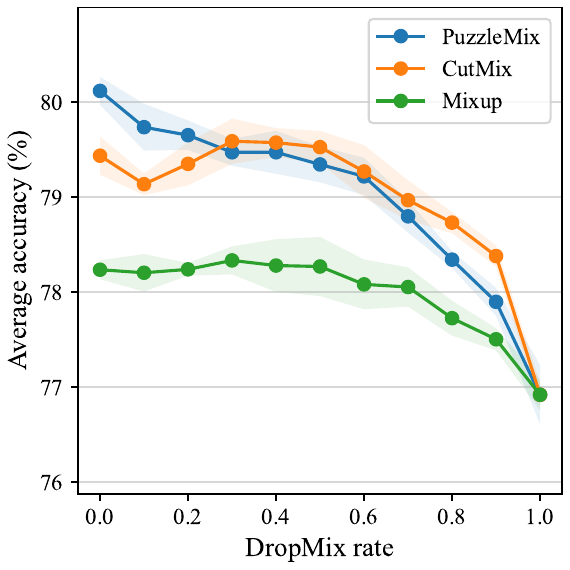}
 \caption{Average accuracy}
 \label{fig:avg_ablation_cifar100_preactrn18}
 \end{subfigure}
 \begin{subfigure}{0.29\linewidth}
 \includegraphics[width=\linewidth]{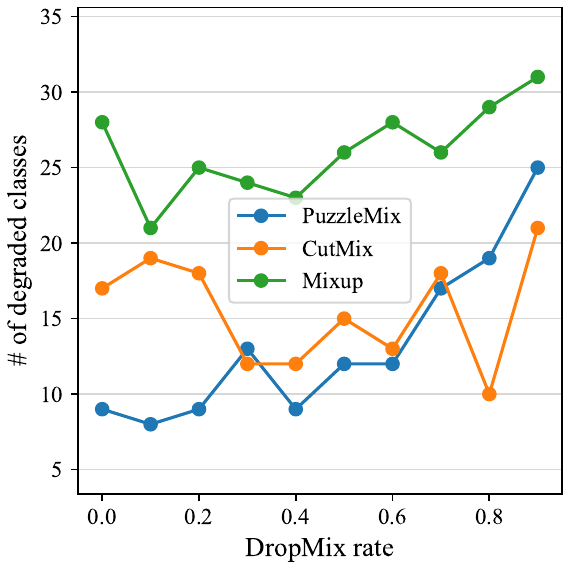}
 \caption{$N_{DC}$}
 \label{fig:deg_ablation_cifar100_preactrn18}
 \end{subfigure}
 \begin{subfigure}{0.29\linewidth}
 \includegraphics[width=\linewidth]{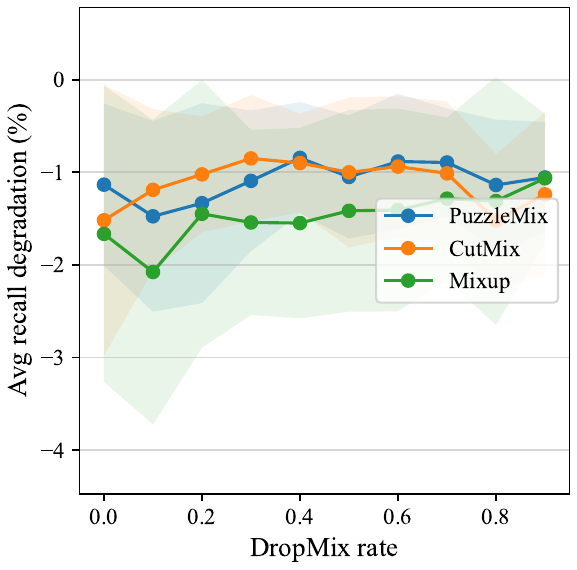}
 \caption{$\overline{\Delta R_{DC}}$}
 \label{fig:avg recall deg_ablation_cifar100_preactrn18}
 \end{subfigure}
 \caption{
 The performances of Mixup (represented in green), CutMix (represented in orange), and PuzzleMix (represented in blue) with DropMix, respectively, are analyzed in terms of class-dependent metrics as a function of the DropMix rate. This analysis was conducted on the PreActRN18 trained on CIFAR-100 dataset. In (a), the average accuracy of the DropMix method surpasses that of the respective MSDA methods (where the DropMix rate is 0) for low DropMix rates. In (b), the number of degraded classes is reduced for a small DropMix rate by DropMix method across all MSDA methods. In (c), the average change in recall for degraded classes is mitigated across the entire range of DropMix.  This figure averages values over five models trained with distinct random seeds, with the shaded area representing the standard deviation of each experiment.
 }
 \label{fig:ablation_cifar100_preactrn18}
\end{figure*}

\begin{figure*}
 \centering
 \begin{subfigure}{0.29\linewidth}
 \includegraphics[width=\linewidth]{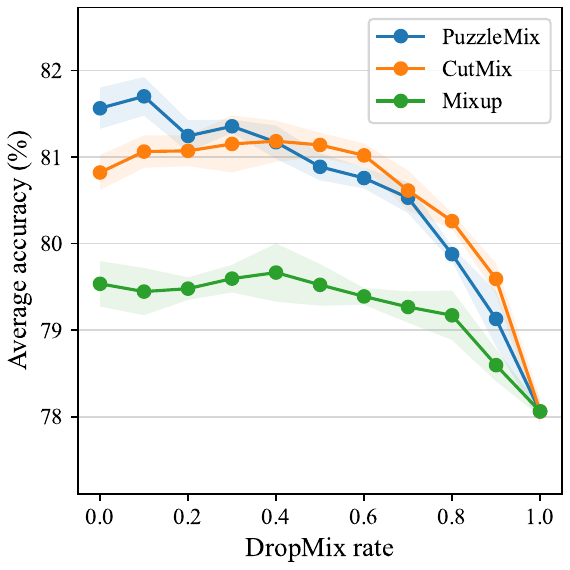}
 \caption{Average accuracy}
 \label{fig:avg_ablation_cifar100_preactrn34}
 \end{subfigure}
 \begin{subfigure}{0.29\linewidth}
 \includegraphics[width=\linewidth]{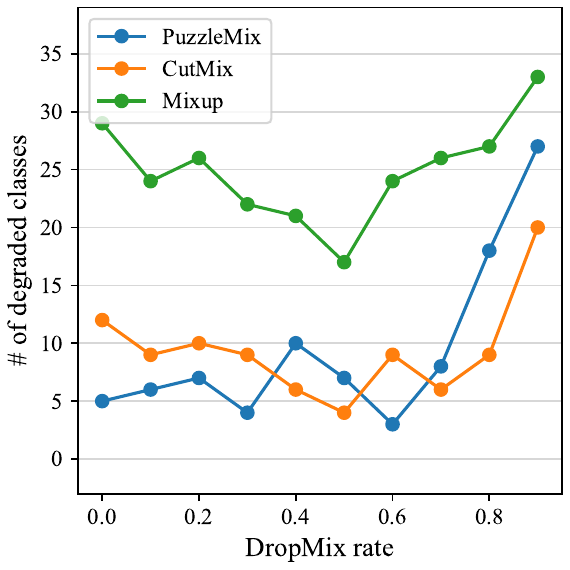}
 \caption{$N_{DC}$}
 \label{fig:deg_ablation_cifar100_preactrn34}
 \end{subfigure}
 \begin{subfigure}{0.29\linewidth}
 \includegraphics[width=\linewidth]{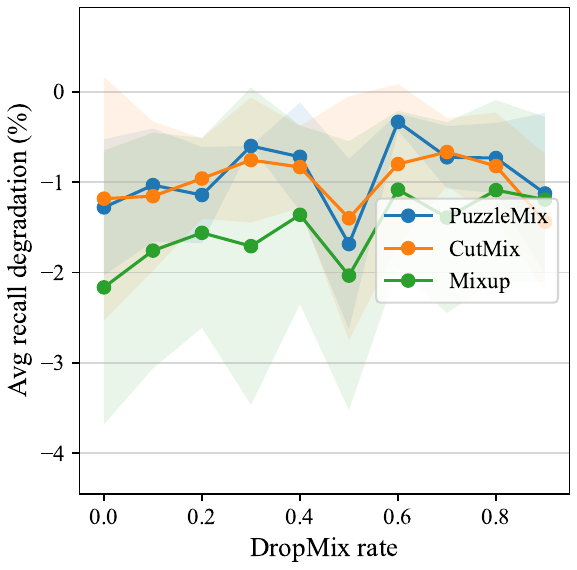}
 \caption{$\overline{\Delta R_{DC}}$}
 \label{fig:avg recall deg_ablation_cifar100_preactrn34}
 \end{subfigure}
 \caption{
 The same analysis with \cref{fig:ablation_cifar100_preactrn18} was conducted on the PreActRN34 trained on CIFAR-100 dataset. This figure averages values over five models trained with distinct random seeds, with the shaded area representing the standard deviation of each experiment.
 }
 \label{fig:ablation_cifar100_preactrn34}
\end{figure*}

\begin{figure*}
 \centering
 \begin{subfigure}{0.29\linewidth}
 \includegraphics[width=\linewidth]{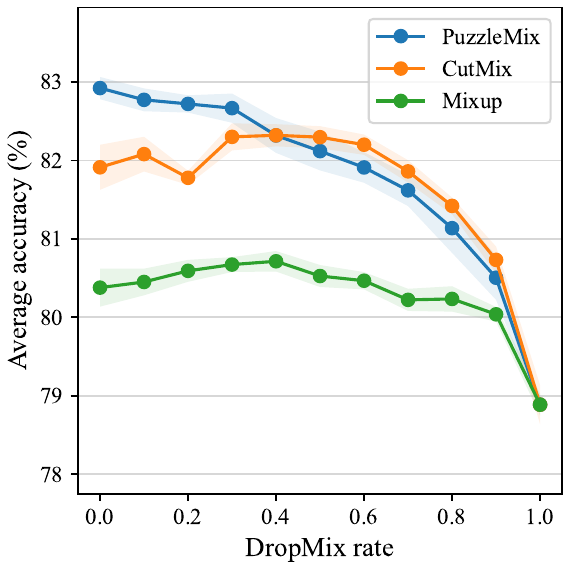}
 \caption{Average accuracy}
 \label{fig:avg_ablation_cifar100_preactrn50}
 \end{subfigure}
 \begin{subfigure}{0.29\linewidth}
 \includegraphics[width=\linewidth]{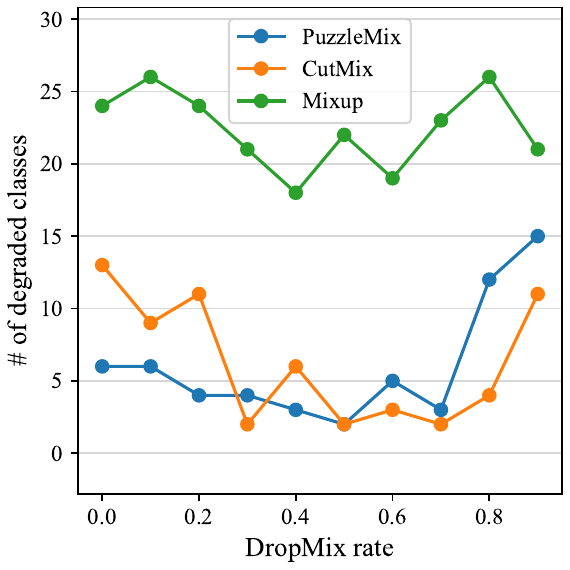}
 \caption{$N_{DC}$}
 \label{fig:deg_ablation_cifar100_preactrn50}
 \end{subfigure}
 \begin{subfigure}{0.29\linewidth}
 \includegraphics[width=\linewidth]{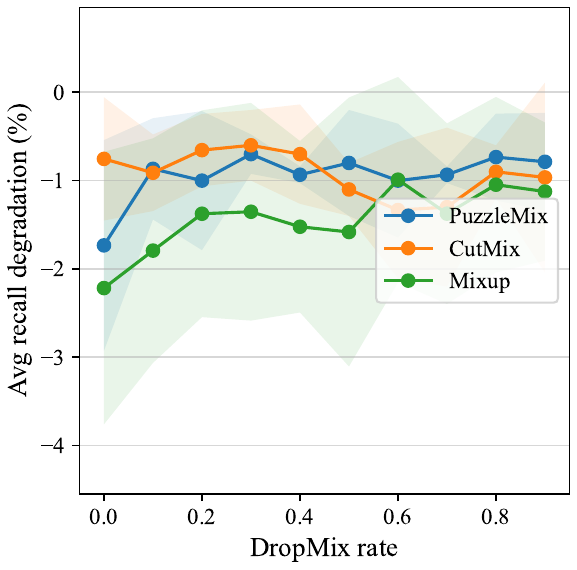}
 \caption{$\overline{\Delta R_{DC}}$}
 \label{fig:avg recall deg_ablation_cifar100_preactrn50}
 \end{subfigure}
 \caption{
 The same analysis with \cref{fig:ablation_cifar100_preactrn18} was conducted on the PreActRN50 trained on CIFAR-100 dataset. This figure averages values over five models trained with distinct random seeds, with the shaded area representing the standard deviation of each experiment.
 }
 \label{fig:ablation_cifar100_preactrn50}
\end{figure*}

\begin{figure*}
 \centering
 \begin{subfigure}{0.42\linewidth}
 \includegraphics[width=\linewidth]{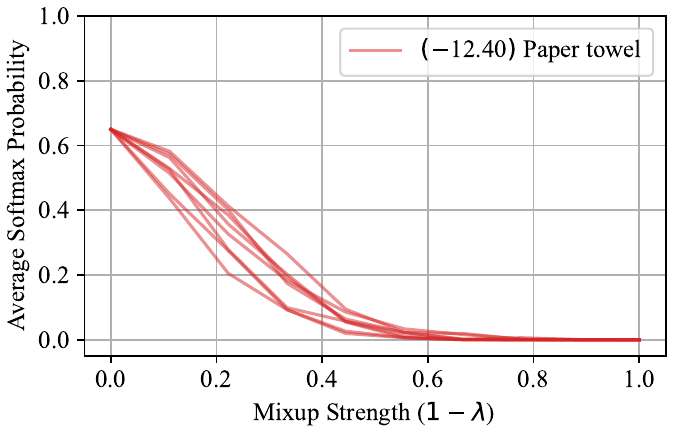}
 \caption{Paper towel}
 \end{subfigure}
  \begin{subfigure}{0.42\linewidth}
 \includegraphics[width=\linewidth]{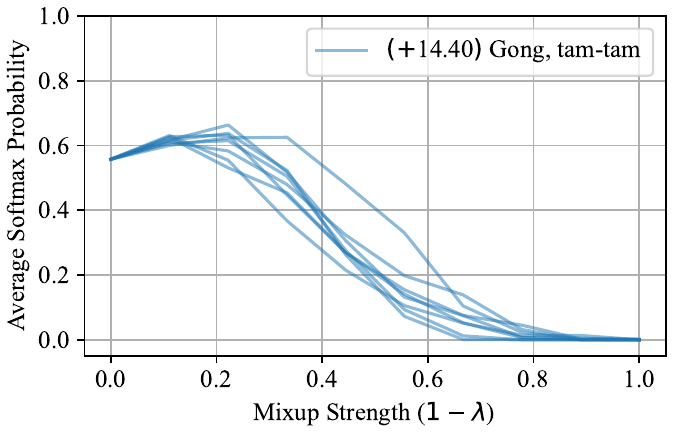}
 \caption{Gong, tam-tam}
 \end{subfigure}
   \begin{subfigure}{0.42\linewidth}
 \includegraphics[width=\linewidth]{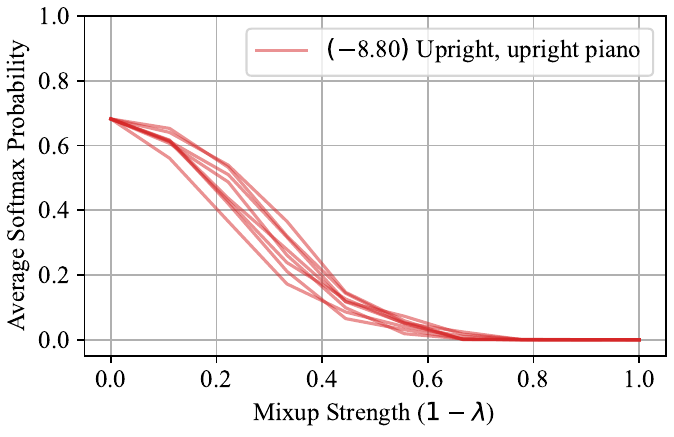}
 \caption{Upright piano}
 \end{subfigure}
  \begin{subfigure}{0.42\linewidth}
 \includegraphics[width=\linewidth]{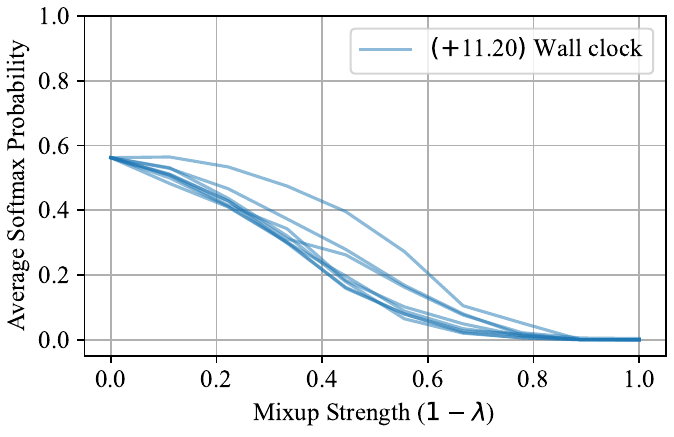}
 \caption{Wall clock}
 \end{subfigure}
  \begin{subfigure}{0.42\linewidth}
 \includegraphics[width=\linewidth]{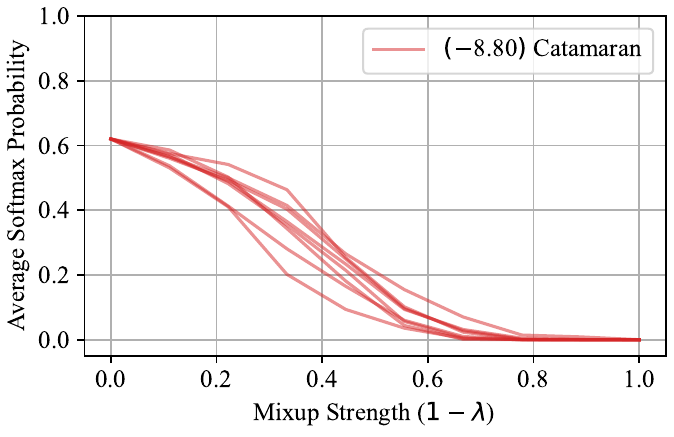}
 \caption{Catamaran}
 \end{subfigure}
  \begin{subfigure}{0.42\linewidth}
 \includegraphics[width=\linewidth]{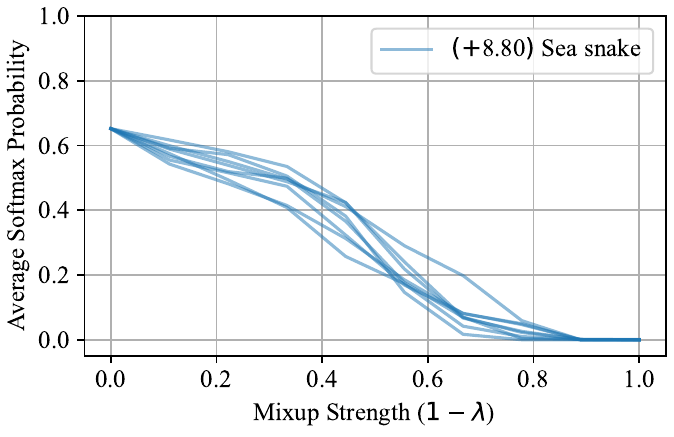}
 \caption{Sea snake}
 \end{subfigure}
  \begin{subfigure}{0.42\linewidth}
 \includegraphics[width=\linewidth]{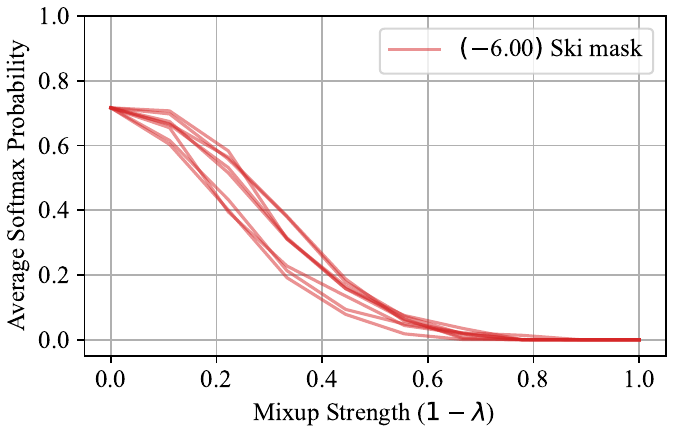}
 \caption{Ski mask}
 \end{subfigure}
  \begin{subfigure}{0.42\linewidth}
 \includegraphics[width=\linewidth]{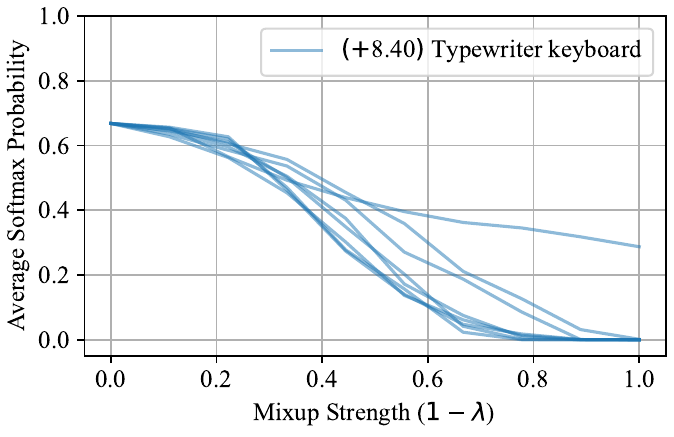}
 \caption{Typewriter keyboard}
 \end{subfigure}
 \caption{
    The impact of Mixup strength on the label information. Each line represents a different class as counterparts in Mixup. The value before the class name indicates the class recall change before and after Mixup, observed in our experiment. As the strength of Mixup increases, the label information for the class is progressively lost. This loss occurs at a noticeably faster rate for the \textcolor{red}{\textit{degraded class}} compared to the \textcolor{blue}{\textit{improved class}}. These findings were obtained from the validation set of ImageNet using the official PyTorch ResNet50 model, which has been pre-trained on the ImageNet dataset.
 }
 \label{fig:impact}
\end{figure*}

\begin{figure*}[t!]
 \centering
 \begin{subfigure}{0.49\linewidth}
 \centering
     \includegraphics[width=0.7\linewidth]{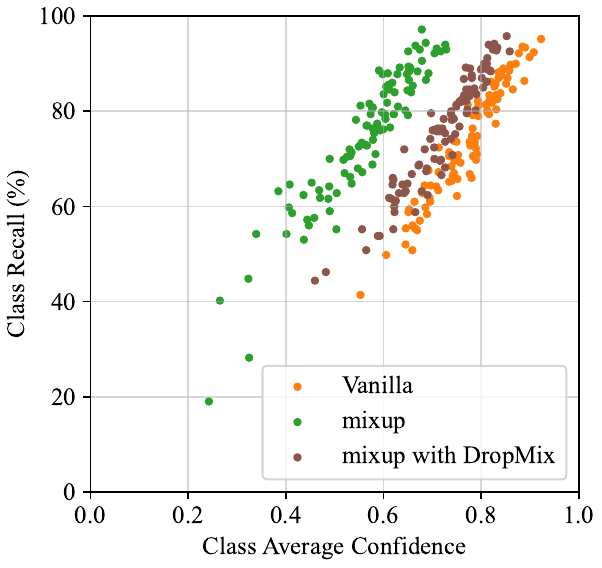}
     \caption{Class confidence v.s. $R$}
     \label{fig:confdence_recall_confidence_recall_wrn28_2_mixup}
 \end{subfigure}
  \begin{subfigure}{0.49\linewidth}
  \centering
     \includegraphics[width=0.68\linewidth]{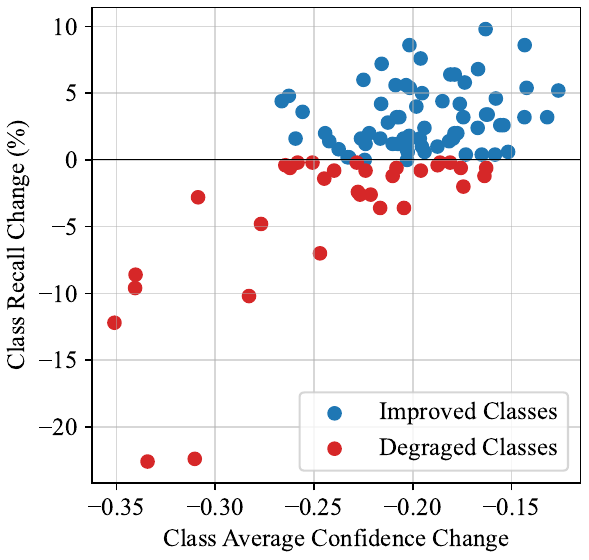}
     \caption{Class confidence change v.s. $\Delta R_{Mixup}$}
     \label{fig:confdence_recall2_change_confidence_recall_wrn28_2_mixup}
 \end{subfigure}
 \caption{
    Correlation between class confidence and recall. Each dot represents an individual class. This analysis was carried out based on the average results from various WideResNet28-2 models trained on the CIFAR-100 dataset.
    In (a), Mixup (represented in green) exhibits a markedly lower confidence distribution compared to the vanilla model (represented in orange). Mixup with DropMix (represented in brown) alleviates the shift in confidence and rescues underperforming classes.
    In (b), most of the classes suffuring sevier recall degradattion also harm from confidence (left bottom). 
    This figure averages values over five models trained with distinct random seeds.
 }
 \label{fig:confidence_recall_wrn28_2_mixup}
\end{figure*}

\begin{figure*}[t!]
 \centering
 \begin{subfigure}{0.49\linewidth}
 \centering
     \includegraphics[width=0.7\linewidth]{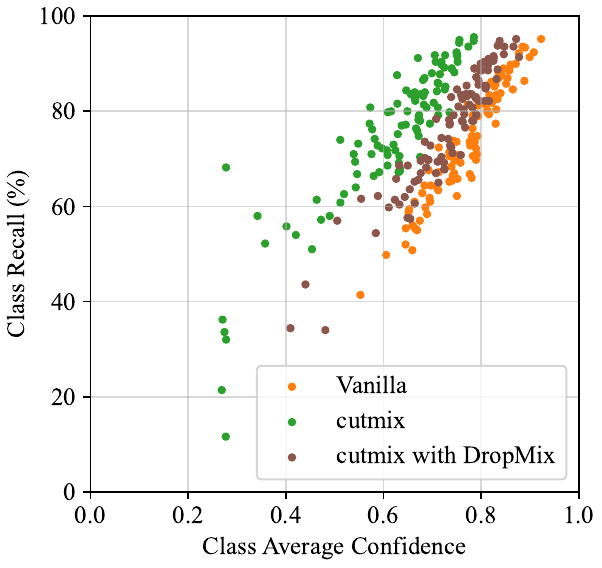}
     \caption{Class confidence v.s. $R$}
     \label{fig:confdence_recall_confidence_recall_wrn28_2_cutmix}
 \end{subfigure}
  \begin{subfigure}{0.49\linewidth}
  \centering
     \includegraphics[width=0.68\linewidth]{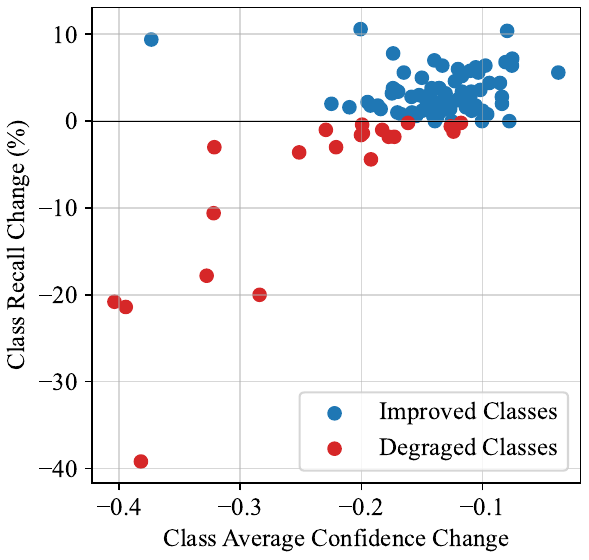}
     \caption{Class confidence change v.s. $\Delta R_{Mixup}$}
     \label{fig:confdence_recall2_change_confidence_recall_wrn28_2_cutmix}
 \end{subfigure}
 \caption{
    The same analysis with \cref{fig:confidence_recall_wrn28_2_mixup} was conducted on the WideResNet28-2 with CutMix.
    This figure averages values over five models trained with distinct random seeds.
 }
 \label{fig:confidence_recall_wrn28_2_cutmix}
\end{figure*}

\begin{figure*}[t!]
 \centering
 \begin{subfigure}{0.49\linewidth}
 \centering
     \includegraphics[width=0.7\linewidth]{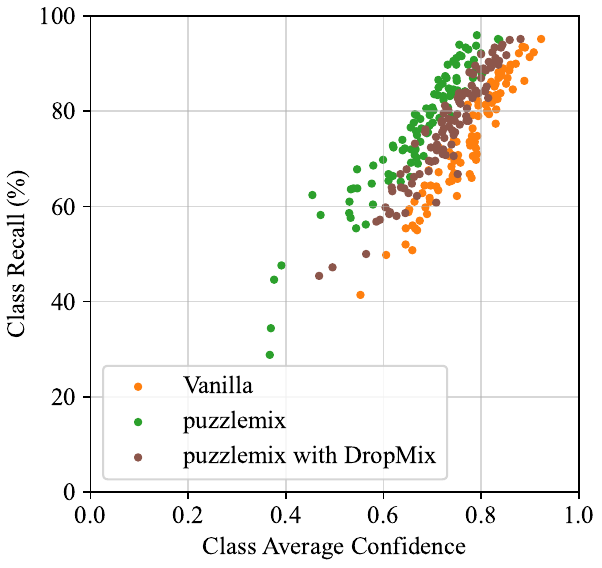}
     \caption{Class confidence v.s. $R$}
     \label{fig:confdence_recall_confidence_recall_wrn28_2_puzzlemix}
 \end{subfigure}
  \begin{subfigure}{0.49\linewidth}
  \centering
     \includegraphics[width=0.68\linewidth]{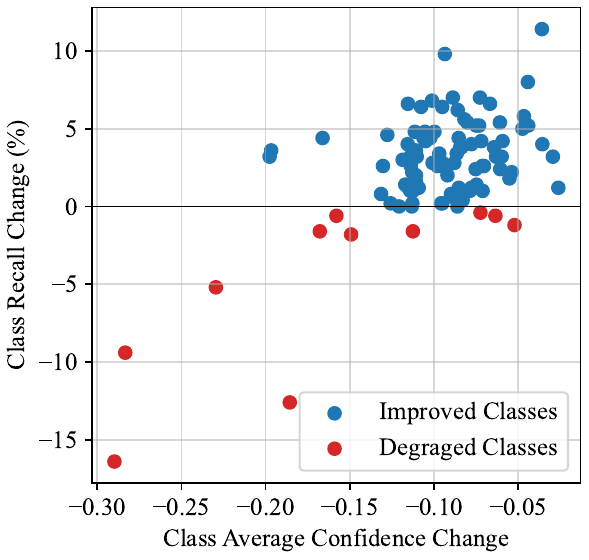}
     \caption{Class confidence change v.s. $\Delta R_{Mixup}$}
     \label{fig:confdence_recall2_change__confidence_recall_wrn28_2_puzzlemix}
 \end{subfigure}
 \caption{
    The same analysis with \cref{fig:confidence_recall_wrn28_2_mixup} was conducted on the WideResNet28-2 with PuzzleMix.
    This figure averages values over five models trained with distinct random seeds.
 }
 \label{fig:confidence_recall_wrn28_2_puzzlemix}
\end{figure*}

Specifically, \cref{fig:cr_ex_mixup,fig:cr_ex_cutmix} illustrates the changes in class recall before and after MSDA in sampled classes on the ImageNet dataset. \Cref{fig:crc_cutmix_imagenet,fig:crc_mixup_imagenet} is a demonstration of class dependency and its mitigation with DropMix on the CutMix-ImageNet setting. \Cref{fig:ablation_cifar100_preactrn18,fig:ablation_cifar100_preactrn34,fig:ablation_cifar100_preactrn50} illustrates the fluctuation in the class dependency metrics as a function of the hyperparameter, DropMix rate, on CIFAR-100 dataset across various networks. 
\Cref{fig:impact} illustrates the impact of Mixup strength on the label information in ImageNet dataset.
\Cref{fig:confidence_recall_wrn28_2_mixup,fig:confidence_recall_wrn28_2_cutmix,fig:confidence_recall_wrn28_2_puzzlemix} illustrates the confidence-recall analysis on CIFAR-100 dataset.

\end{document}